\documentclass[letterpaper]{article} %
\usepackage[preprint]{aaai2027}  %
\usepackage[hyphens]{url}  %
\usepackage{graphicx} %
\usepackage{natbib}  %
\usepackage{caption} %
\usepackage{booktabs}
\usepackage{amsmath}
\usepackage{amssymb}
\usepackage{multirow}

\newlength{\hdrvshift}
\title{SleepVLM: A Rule-Grounded Vision-Language Model for Auditable Sleep Staging}

\author{
    Guifeng Deng\textsuperscript{\rm 1,3},
    Pan Wang\textsuperscript{\rm 2},
    Mengfan Niu\textsuperscript{\rm 1},
    Jiquan Wang\textsuperscript{\rm 4},
    Shuying Rao\textsuperscript{\rm 1,3},
    Junyi Xie\textsuperscript{\rm 1},
    Xi'ang Chen\textsuperscript{\rm 1,3},
    Sha Zhao\textsuperscript{\rm 4},
    Gang Pan\textsuperscript{\rm 4},
    Wanjun Guo\textsuperscript{\rm 1,4,5},
    Tao Li\textsuperscript{\rm 1,4,5,*},
    Haiteng Jiang\textsuperscript{\rm 1,2,4,5,*}
}
\affiliations{
    \textsuperscript{\rm 1}Affiliated Mental Health Center \& Hangzhou Seventh
    People's Hospital, School of Brain Science and Brain Medicine, and Liangzhu
    Laboratory, Zhejiang University School of Medicine, Hangzhou 310058, China\\
    \textsuperscript{\rm 2}Department of Psychiatry and Mental Health, Wenzhou
    Medical University, Wenzhou 325035, Zhejiang Province, China\\
    \textsuperscript{\rm 3}College of Biomedical Engineering \& Instrument
    Science, Zhejiang University, Hangzhou 310058, China\\
    \textsuperscript{\rm 4}MOE Frontier Science Center for Brain Science and
    Brain-machine Integration, State Key Laboratory of Brain-machine
    Intelligence, Zhejiang University, Hangzhou 311121, China\\
    \textsuperscript{\rm 5}Zhejiang Key Laboratory of Clinical and Basic
    Research for Psychiatric Diseases, Hangzhou 310058, China\\
    \textsuperscript{\rm *}Correspondence: litaozjusc@zju.edu.cn (T.L.),
    h.jiang@zju.edu.cn (H.J.)
}

\begin{document}
\maketitle
\pagestyle{plain}\thispagestyle{plain}

\begin{abstract}
Sleep staging is essential for sleep assessment and disorder diagnosis. In recent years, automatic sleep staging systems have achieved accuracy approaching that of human experts, but the black-box nature of their predictions hinders clinical adoption. Existing interpretability methods offer partial insight into model behavior, but their outputs still require expert reinterpretation and do not provide a direct basis for auditing individual predictions. To improve trustworthiness, we propose the task of auditable sleep staging. To solve this task, we present SleepVLM, a vision-language model that casts sleep staging as visual reasoning over rendered polysomnography (PSG) waveform images. For each epoch, SleepVLM outputs a stage together with the applicable American Academy of Sleep Medicine (AASM) rules and an auditable rationale. The model is trained using a two-stage framework: Waveform-Perceptual Pre-training followed by Rule-Grounded Supervised Fine-tuning over a mixture of fine-grained and coarse annotations. Experiments on four datasets show that SleepVLM outperforms state-of-the-art methods on average. An automated AASM-feature audit shows broad coverage of stage-defining evidence in the rationales, and independent experts validate their reasoning quality. To facilitate further research, we construct and release MASS-EX, an expert-annotated dataset for rule-grounded sleep staging with AASM rule annotations and expert-written rationales.
\end{abstract}

\section{Introduction}
\begin{figure*}[!t]
  \centering
  \includegraphics[width=\textwidth]{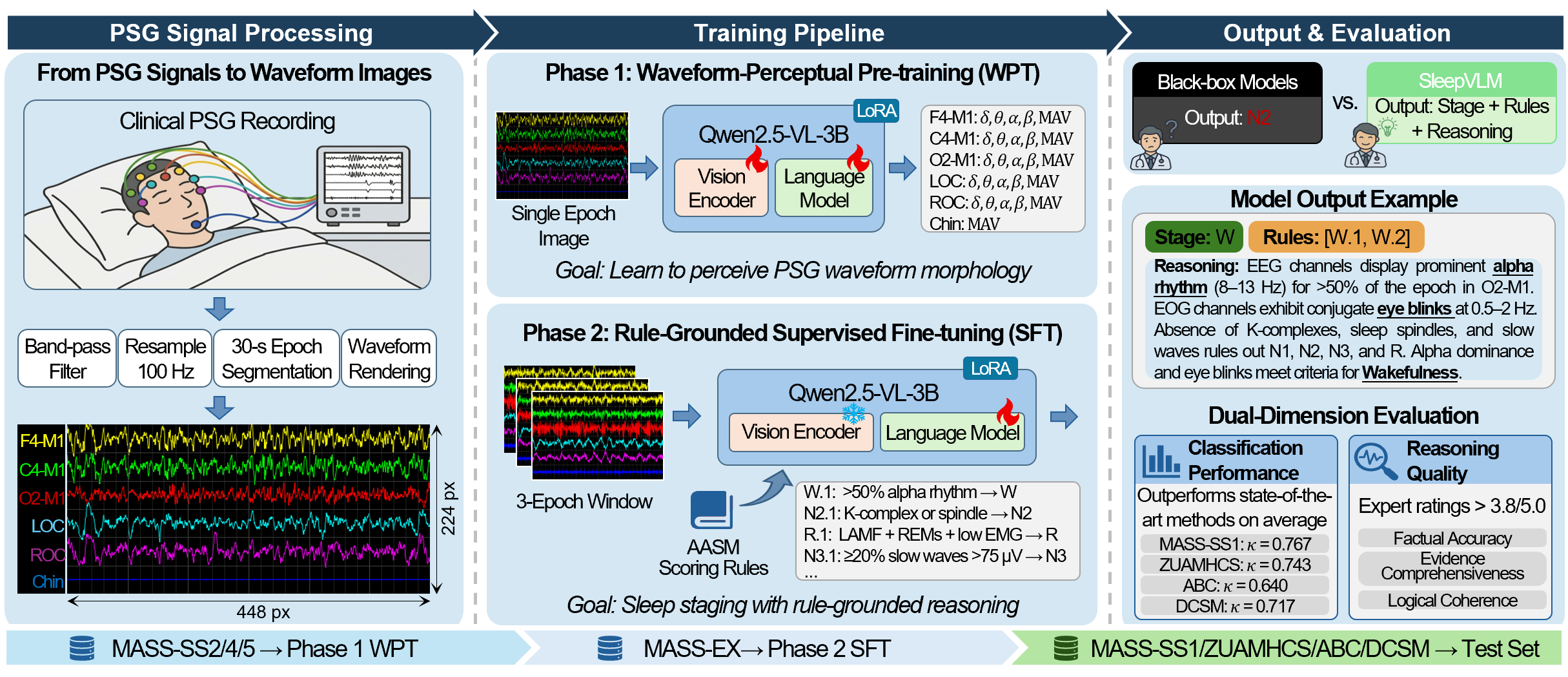}
  \caption{Overview of SleepVLM.}
  \label{fig:pipeline}
\end{figure*}

Automatic sleep staging is essential for sleep assessment and disorder diagnosis. In clinical practice, sleep stages are scored from a polysomnogram (PSG), which records multiple physiological signals such as the electroencephalogram (EEG), the electrooculogram (EOG) and the electromyogram (EMG). The PSG is segmented into 30-second epochs, and each epoch is scored by a sleep expert as wakefulness (W), non-rapid eye movement (NREM) stages N1, N2 and N3, or rapid eye movement sleep (R), according to the American Academy of Sleep Medicine (AASM) scoring criteria \cite{berry2023aasm}. This manual process is labor-intensive and time-consuming, and the agreement between scorers is limited \cite{lee2022interrater,rosenberg2013american}.

In recent years, many deep learning models have been proposed for automatic sleep staging, including convolutional and recurrent models such as DeepSleepNet and TinySleepNet \cite{supratak2017deepSleepNet,supratak2020tinysleepnet}, attention-based models such as AttnSleep and SleepTransformer \cite{eldele2021attnSleep,phan2022sleeptransformer}, and fully convolutional models such as U-Time and U-Sleep \cite{perslev2019uTime,perslev2021uSleep}. Several of them have achieved accuracy approaching that of human experts \cite{phang2025explainableMTCNN,vanderdonckt2023traditionalML}. However, they are all black-box classifiers and output only a stage label. A clinician cannot tell which waveform features lead to a prediction, or whether the prediction follows the AASM rules that the clinician has to apply. This lack of interpretability hinders clinical adoption \cite{phan2022sleeptransformer,horie2022sleepcam,park2025distillsleep,muto2023looking}.

Several methods add interpretability to sleep staging, including attention visualization \cite{phan2022sleeptransformer}, saliency mapping obtained by gradient attribution or relevance propagation \cite{dutt2023sleepxai,vaquerizo2023explainablechildren,lee2025sleepXViT}, learned prototypes \cite{pei2025wavesleepnet} and architectural designs that inject domain knowledge into the classifier \cite{niknazar2024multilevel,alhussaini2022serf}. These methods offer partial insight into model behavior. However, they indicate where the model attends rather than why an epoch should be scored as a given stage under the AASM rules, so their outputs still require expert reinterpretation. Clinicians instead need an explanation in their own diagnostic terms that makes each staging decision auditable against the rules they apply themselves \cite{bienefeld2023xai}. We call this task auditable sleep staging.

Vision-language models (VLMs) offer another way to approach this problem. In medical imaging, VLMs have been used to generate diagnostic reports and to support interactive analysis of images \cite{Holland2025RetinaVLM,Lu2024PathChat,Christensen2024EchoCLIP}. However, a high accuracy does not guarantee a reliable explanation. Among the medical image questions that GPT-4V answered correctly, 35.5\% were supported by a flawed rationale, and errors in image comprehension were the most frequent cause \cite{Jin2024HiddenFlaws}. A VLM for sleep staging therefore needs domain-specific training to read waveforms correctly, and its explanations have to be evaluated separately from its classification accuracy.

In this paper, we propose SleepVLM (Figure~\ref{fig:pipeline}), which mirrors the workflow of a sleep technologist. SleepVLM renders six PSG channels as a waveform image and takes the preceding, current and subsequent epochs as input, with the AASM rules written into the system prompt during training. For each epoch, the model outputs a sleep stage, the identifiers of the applicable rules, and an auditable rationale that describes the stage-defining features it observes. Different from post-hoc explanation methods, SleepVLM produces the prediction and its rationale at the same time, so the rationale can be checked directly against the AASM manual.

Our contributions are as follows:
\begin{itemize}
\item We propose the task of auditable sleep staging and present SleepVLM, which casts sleep staging as visual reasoning over rendered PSG waveform images. To our knowledge, SleepVLM is the first vision-language model to ground sleep-staging decisions in AASM rules.
\item We propose a two-stage training framework: Waveform-Perceptual Pre-training on per-second spectral and amplitude targets, followed by Rule-Grounded Supervised Fine-tuning over a mixture of fine-grained and coarse annotations. Ablation studies verify the contribution of each component.
\item SleepVLM is validated on four datasets, outperforming state-of-the-art methods on average and generating high-quality auditable rationales.
\item We construct and release MASS-EX, to our knowledge the first expert-annotated dataset for rule-grounded sleep staging, covering 62 subjects and 59,317 epochs with AASM rule annotations and expert-written rationales for a subset.
\end{itemize}

\section{Related Work}

\subsection{Automatic Sleep Staging}
Automatic sleep staging has been driven by deep learning, and its accuracy now approaches that of human experts. Early models combined a CNN for features within an epoch with a recurrent layer for context across epochs \cite{supratak2017deepSleepNet,supratak2020tinysleepnet}, while \citet{phan2019seqSleepNet} built a hierarchical RNN over time-frequency images. \citet{qu2020resnetmha,eldele2021attnSleep,phan2022sleeptransformer} replaced the recurrent layers with multi-head self-attention, the first over EEG decomposed into AASM frequency bands. Others changed the input or the unit of prediction: \citet{phan2022xSleepNet} added the raw signal to the time-frequency image, \citet{perslev2019uTime,perslev2021uSleep} segmented a whole recording in one pass, and \citet{jia2021salientSleepNet} separated the EEG and EOG streams. Later work targeted unseen cohorts, through arbitrary montages \cite{guillot2021robustSleepNet}, feature alignment \cite{wang2024sleepDG}, and multi-center pretraining \cite{deng2024lpsgm}, which reached expert-level agreement in a prospective study. Such models are now retrained and evaluated under a unified protocol \cite{deirossi2026sleepyland}.

\subsection{Interpretable Sleep Staging}
The black-box nature of these models hinders their clinical adoption \cite{phan2022sleeptransformer,horie2022sleepcam}. Several methods make the evidence behind a prediction explicit. One group marks where in the signal a decision comes from. \citet{phan2022sleeptransformer} used self-attention to show which parts of an epoch and which neighboring epochs matter. \citet{dutt2023sleepxai} and \citet{vaquerizo2023explainablechildren} applied Grad-CAM after training, and \citet{horie2022sleepcam} used class activation mapping. \citet{lee2025sleepXViT} proposed SleepXViT, which propagates relevance through a vision transformer over rendered PSG images, the same form of input SleepVLM takes. A second group names the unit a decision relies on. \citet{pei2025wavesleepnet} learned wave prototypes and scored an epoch by their presence and proportion. \citet{niknazar2024multilevel} constrained the first convolutional layer to Gabor kernels, so the learned waveforms can be read after training. \citet{alhussaini2022serf} projected deep embeddings onto features derived from the AASM manual and classified them with a simple classifier. A third group avoided deep models altogether \cite{mai2025aisleep,vanderdonckt2023traditionalML}. Two studies measured how much of this evidence is recovered. \citet{horie2022sleepcam} found that the highlighted intervals cover alpha rhythm, eye movements, spindles and slow waves, but not K complexes or arousals. \citet{lee2025sleepXViT} reported that their heatmaps can be read under the AASM rules for W, N1 and N2, while those for N3 and R require further discussion. These methods output a location, a weight or a feature value rather than the rule that the epoch satisfies, so the scorer still has to make this connection. \citet{hardarson2026staging} removed this step by executing the AASM rules as code and generating a justification from the trace, but reached only 60.5\% agreement ($\kappa=0.42$) against a ten-scorer consensus, far below the methods above.

\subsection{Language Models for Physiological Signals}
Vision-language models offer a new way to explain a prediction in natural language. In medical imaging, \citet{Lu2024CONCH} aligned histopathology images with text for retrieval and captioning, and \citet{ffagpt2024} generated fundus angiography reports and answered questions about them. Two lines of work apply this idea to physiological signals. \citet{xu2026sleeplm} proposed SleepLM, which aligns raw multimodal PSG with text over a paired dataset of more than 100,000 hours from over 10,000 subjects. The other line renders the signal as an image. \citet{gucerebragloss} instruction-tuned a vision-language model on clinical EEG images, using a waveform detector and a teacher model to produce the descriptions. \citet{qiu2025eegvlm} aligned a visual encoder with CLIP features and distilled chain-of-thought supervision from a larger model. In all three methods the text used for supervision is generated by templates or by another model, so a sentence cannot be checked against an external standard, and none of them reports which criterion it applied. In contrast, SleepVLM uses the AASM manual as this standard. For every epoch it reports the identifiers of the applicable rules with the rationale, and evaluates the rationale separately from the stage.

\section{MASS-EX Dataset}

Public sleep datasets label the stage of each epoch, but not the AASM rules it follows from. To support rule-grounded supervision, we construct and release MASS-EX, an expert-annotated dataset based on all 62 MASS-SS3 subjects and their 59,317 original epochs \cite{oreilly2014mass}. The underlying rule library comprises 15 adult sleep staging rules derived from Version 3 of the AASM scoring manual \cite{berry2023aasm} and operationalized for the six PSG channels used in this work. Rule N3.1, for instance, is met when slow wave activity (0.5--2.0 Hz, peak-to-peak amplitude above 75 $\mu$V) makes up at least 20\% of the epoch, predominantly in the frontal derivations. The rule set was developed jointly by a trained sleep technologist and a senior sleep medicine physician with over a decade of clinical experience. The complete rule set is given in the Appendix.

Because the model input uses a preceding--current--subsequent three-epoch window, the first and last epoch of each recording cannot serve as the center epoch and are excluded from annotation. This yields 59,193 annotated central epochs. Among them, 5,006 epochs from five subjects carry fine-grained annotations consisting of applicable rule identifiers, an expert-written rationale, and a sleep stage label. The remaining 54,187 epochs from 57 subjects carry coarse annotations consisting of applicable rule identifiers and a sleep stage label only. Annotations were produced through an expert-driven, machine-assisted pipeline: the two experts first authored high-quality exemplar annotations for each sleep stage; a locally deployed Qwen2.5-VL-72B-Instruct model \cite{bai2025qwen25vl} then generated draft annotations for all target epochs using these exemplars as few-shot demonstrations; the sleep technologist manually reviewed and corrected every generated annotation; and the senior physician independently verified and finalized the results. The full pipeline is given in the Appendix.

\section{Method}

SleepVLM takes the rendered waveform images of three consecutive epochs and outputs a sleep stage, the identifiers of the applicable rules, and an auditable rationale for the central epoch. As shown in Figure~\ref{fig:pipeline}, we obtain this behavior by adapting a pretrained vision-language model in two stages: waveform-perceptual pre-training (WPT), followed by rule-grounded supervised fine-tuning (SFT). In both stages we give the model an image sequence $I$ and a system prompt $p$, and train it to generate a target text $y$:
\begin{equation}
\mathcal{L}(\theta) = -\sum\nolimits_{t} \log P_{\theta}\!\left(y_{t} \mid y_{<t}, I, p\right),
\end{equation}
where $\theta$ denotes the trainable parameters. The two stages differ in what $I$, $p$ and $y$ contain, and in which parameters we train. We use low-rank adaptation (LoRA) \cite{hu2022lora} for parameter-efficient fine-tuning in both stages.

\subsection{PSG Signal Processing and Waveform Rendering}

We render each epoch as a standardized multichannel waveform image, so that the model can inspect PSG traces as a technologist does. We use six channels, following the AASM-recommended montage for adult sleep staging: three EEG derivations (F4-M1, C4-M1 and O2-M1), two EOG channels (LOC and ROC), and chin EMG. EEG and EOG are band-pass filtered at 0.3--35 Hz and chin EMG at 10--100 Hz, both with fourth-order zero-phase Butterworth filters. All channels are then resampled to 100 Hz and segmented into non-overlapping 30-second epochs. Each epoch is therefore a six-channel signal $X \in \mathbb{R}^{6 \times 3000}$.

Each image is $448 \times 224$ pixels with a black background. The six channels are stacked vertically in a fixed order, each drawn in its own color. Each channel also has a fixed amplitude scale: $\pm 50\,\mu$V for EEG and EOG, and $\pm 40\,\mu$V for chin EMG. Time grid lines are drawn at 1-second and 5-second intervals. We state the same channel order, colors and amplitude scales in $p$. The model can therefore tell which channel a waveform belongs to, and read amplitudes in microvolts. The AASM amplitude criteria are stated in the same unit. Signal excursions beyond a channel boundary are not clipped, so a waveform that extends into an adjacent channel stays visible and indicates an extreme amplitude.

\subsection{Waveform-Perceptual Pre-training}

WPT is a perceptual pretext task. It teaches the model to interpret waveform images quantitatively, before any staging supervision is introduced. Here $I$ is a single 30-second waveform image, and $p$ states the rendering parameters and the task. The target $y$ is a set of per-second spectral and amplitude descriptors rather than a sleep stage. For each EEG or EOG channel and each one-second window, $y$ comprises the delta, theta, alpha and beta band powers in decibels, together with the mean absolute value (MAV) in microvolts. For chin EMG it comprises the MAV alone. Band powers are obtained by Welch spectral estimation over each window. Every element of $y$ is computed from the signal itself, so this stage needs no expert annotation. In WPT the vision encoder is unfrozen, which lets the model adapt to the PSG waveform image domain. Here $\theta$ covers the encoder together with the LoRA parameters.

\subsection{Rule-Grounded Supervised Fine-tuning}

SFT adapts the model to perform sleep staging with structured, rule-cited reasoning. Here $I$ is three consecutive epoch images, representing the preceding, current and subsequent epochs, and the staging decision is always made for the central one. Several AASM rules refer to the epoch before or after the central epoch, so both neighbors are needed.

We ground staging decisions in the AASM rules through $p$. All 15 rules are injected into $p$ in full, together with the role and task definition, the rendering parameters and the output format. The task instructions require the model to report the waveform features of the central epoch, with their channel, timing, morphology, frequency and amplitude. The model must also cite the rule identifiers that support its decision, and combine these findings into a rationale. The target $y$ therefore has three parts: the stage, the applicable rule identifiers, and the rationale.

We mix the fine-grained and coarse annotations of MASS-EX during training. The fine-grained annotations supervise all three parts of $y$. The coarse annotations supervise the stage and the identifiers only, and we adjust the task instructions and the output format in $p$ accordingly. At inference the model always produces the full $y$, and generates a complete rationale regardless of how the training sample was supervised. In SFT the vision encoder is frozen, which preserves the visual representations learned in WPT. Here $\theta$ covers only the LoRA parameters of the language model. The complete prompt and both output templates are given in the Appendix.

\section{Experiments}
\subsection{Datasets and Settings}

\begin{table}[t]
\centering\fontsize{7.8}{9.3}\selectfont
\setlength{\tabcolsep}{2.5pt}
\caption{Data used for model development and evaluation.}
\label{tab:data}
\begin{tabular}{lccl}
\toprule
Dataset & Subjects & Sampling rate (Hz) & Use \\
\midrule
\multicolumn{4}{l}{\textit{Development}} \\
MASS-SS2/SS4/SS5 & 85  & 256 & WPT \\
MASS-EX          & 5   & 256 & SFT (fine-grained) \\
MASS-EX          & 45  & 256 & SFT (coarse) \\
MASS-EX          & 12  & 256 & Validation \\
\midrule
\multicolumn{4}{l}{\textit{Evaluation}} \\
MASS-SS1         & 53  & 256 & Held-out test \\
ZUAMHCS          & 100 & 512 & External test \\
ABC              & 132 & 256 & External test \\
DCSM             & 255 & 256 & External test \\
\bottomrule
\end{tabular}
\end{table}

The data used in this work are summarized in Table~\ref{tab:data}. All development data come from MASS. MASS-SS2, SS4 and SS5 were used for WPT, which needs no annotation because its supervision targets are computed from the signals. MASS-EX was used for SFT. It annotates the 62 subjects of MASS-SS3, of which 5 carry fine-grained annotations and 57 carry coarse annotations. Of these 57 subjects, 45 were used for training and 12 formed the validation set. MASS-SS1 was the held-out test set. The other three datasets were used for external evaluation: ZUAMHCS is a private clinical dataset, while ABC \cite{bakker2018gastric,zhang2018national} and DCSM \cite{perslev2021uSleep} are public. All splits are subject-independent, so no subject appears in both training and evaluation.

The backbone model is Qwen2.5-VL-3B-Instruct \cite{bai2025qwen25vl}, adapted with LoRA (rank 16, scaling factor 32, dropout 0.05). Both stages were trained with AdamW ($\beta_1 = 0.9$, $\beta_2 = 0.95$, $\epsilon = 10^{-8}$) at a learning rate of $1 \times 10^{-4}$, warmed up linearly over the first 3\% of the steps and then decayed linearly, with weight decay 0.1 and gradient clipping at 1.0. WPT ran for 2 training epochs at an effective batch size of 32 and SFT for 15 training epochs at an effective batch size of 48, with a checkpoint saved every 1,000 steps and the highest-$\kappa$ checkpoint on the validation set retained. Training used bfloat16 mixed precision on eight NVIDIA A100 80\,GB GPUs. Inference ran through vLLM with temperature $10^{-6}$, top-$p$ 0.8 and a maximum of 1,024 new tokens. Classification performance is evaluated with accuracy (Acc), macro-F1 (MF1) and Cohen's kappa ($\kappa$).

\subsection{Classification Performance}
\begin{table*}[t]
\centering\fontsize{8.2}{9.8}\selectfont
\setlength{\tabcolsep}{2.2pt}
\caption{Sleep staging performance on four datasets. Best value per column in bold.}
\label{tab:main}
\begin{tabular}{cc ccc ccc ccc ccc ccc}
\toprule
\multirow{2}{*}[-\hdrvshift]{Method} & \multirow{2}{*}[-\hdrvshift]{Modality} & \multicolumn{3}{c}{MASS-SS1} & \multicolumn{3}{c}{ZUAMHCS} & \multicolumn{3}{c}{ABC} & \multicolumn{3}{c}{DCSM} & \multicolumn{3}{c}{Avg} \\
\cmidrule(lr){3-5}\cmidrule(lr){6-8}\cmidrule(lr){9-11}\cmidrule(lr){12-14}\cmidrule(lr){15-17}
 & & Acc & MF1 & $\kappa$ & Acc & MF1 & $\kappa$ & Acc & MF1 & $\kappa$ & Acc & MF1 & $\kappa$ & Acc & MF1 & $\kappa$ \\
\midrule
AttnSleep       & Signal & 0.801 & 0.756 & 0.723 & 0.752 & 0.701 & 0.666 & 0.705 & 0.674 & 0.602 & 0.662 & 0.600 & 0.548 & 0.730 & 0.683 & 0.635 \\
DeepSleepNet    & Signal & 0.803 & 0.751 & 0.725 & 0.749 & 0.701 & 0.658 & 0.652 & 0.606 & 0.534 & 0.660 & 0.594 & 0.544 & 0.716 & 0.663 & 0.615 \\
LPSGM           & Signal & 0.831 & \textbf{0.794} & 0.763 & \textbf{0.818} & \textbf{0.775} & \textbf{0.750} & 0.683 & 0.663 & 0.580 & 0.657 & 0.616 & 0.552 & 0.747 & 0.712 & 0.661 \\
ResnetMHA       & Signal & 0.784 & 0.761 & 0.709 & 0.751 & 0.712 & 0.667 & 0.716 & \textbf{0.715} & 0.627 & 0.667 & 0.633 & 0.567 & 0.730 & 0.705 & 0.643 \\
RobustSleepNet  & Signal & 0.815 & 0.755 & 0.738 & 0.799 & 0.757 & 0.725 & 0.597 & 0.524 & 0.473 & 0.682 & 0.627 & 0.578 & 0.723 & 0.666 & 0.628 \\
SalientSleepNet & Signal & 0.787 & 0.722 & 0.699 & 0.778 & 0.732 & 0.693 & 0.639 & 0.564 & 0.518 & 0.632 & 0.561 & 0.513 & 0.709 & 0.645 & 0.606 \\
SeqSleepNet     & Signal & 0.799 & 0.742 & 0.712 & 0.812 & 0.757 & 0.737 & 0.669 & 0.623 & 0.546 & 0.663 & 0.601 & 0.549 & 0.736 & 0.681 & 0.636 \\
SleepDG         & Signal & 0.812 & 0.762 & 0.735 & 0.798 & 0.749 & 0.719 & 0.729 & 0.698 & 0.631 & 0.692 & 0.631 & 0.587 & 0.758 & 0.710 & 0.668 \\
TinySleepNet    & Signal & 0.829 & 0.779 & 0.760 & 0.786 & 0.741 & 0.712 & 0.719 & 0.686 & 0.619 & 0.677 & 0.616 & 0.567 & 0.753 & 0.706 & 0.665 \\
U-Sleep         & Signal & 0.812 & 0.749 & 0.732 & 0.781 & 0.720 & 0.697 & 0.720 & 0.675 & 0.614 & 0.686 & 0.620 & 0.576 & 0.750 & 0.691 & 0.655 \\
U-Time          & Signal & 0.770 & 0.703 & 0.675 & 0.776 & 0.715 & 0.691 & 0.682 & 0.642 & 0.562 & 0.663 & 0.587 & 0.539 & 0.723 & 0.662 & 0.617 \\
XSleepNet       & Signal & 0.780 & 0.711 & 0.689 & 0.727 & 0.661 & 0.634 & 0.489 & 0.436 & 0.320 & 0.652 & 0.563 & 0.532 & 0.662 & 0.593 & 0.544 \\
SleepXViT       & Image  & \textbf{0.838} & 0.793 & \textbf{0.771} & 0.776 & 0.724 & 0.694 & 0.703 & 0.671 & 0.592 & 0.786 & 0.728 & 0.712 & 0.776 & 0.729 & 0.692 \\
Resnet-18       & Image  & 0.830 & 0.778 & 0.758 & 0.793 & 0.747 & 0.717 & 0.713 & 0.689 & 0.609 & 0.781 & 0.716 & 0.705 & 0.779 & 0.732 & 0.697 \\
\midrule
\textbf{SleepVLM (ours)} & Image & 0.835 & 0.793 & 0.767 & 0.812 & 0.766 & 0.743 & \textbf{0.739} & 0.706 & \textbf{0.640} & \textbf{0.789} & \textbf{0.733} & \textbf{0.717} & \textbf{0.794} & \textbf{0.750} & \textbf{0.717} \\
\bottomrule
\end{tabular}
\end{table*}
SleepVLM was compared against the 14 baselines listed in Table~\ref{tab:main}: 12 signal-based methods, namely AttnSleep \cite{eldele2021attnSleep}, DeepSleepNet \cite{supratak2017deepSleepNet}, LPSGM \cite{deng2024lpsgm}, ResnetMHA \cite{qu2020resnetmha}, RobustSleepNet \cite{guillot2021robustSleepNet}, SalientSleepNet \cite{jia2021salientSleepNet}, SeqSleepNet \cite{phan2019seqSleepNet}, SleepDG \cite{wang2024sleepDG}, TinySleepNet \cite{supratak2020tinysleepnet}, U-Sleep \cite{perslev2021uSleep}, U-Time \cite{perslev2019uTime} and XSleepNet \cite{phan2022xSleepNet}, together with two image-based methods, SleepXViT \cite{lee2025sleepXViT} and ResNet-18 \cite{he2016resnet}, which take the same rendered waveform images as SleepVLM. All baselines were reimplemented from their publicly released code and retrained on the same data splits and channel configuration, with their published hyperparameters.

SleepVLM attains the highest average accuracy, macro-F1 and $\kappa$ over the four datasets (Table~\ref{tab:main}). On MASS-SS1 and on ZUAMHCS it ranks second in $\kappa$, behind SleepXViT by 0.004 and behind LPSGM by 0.007. Across the three metrics on all four datasets, it never trails the best method by more than 0.009.

Neither leader retains its advantage on the remaining datasets. Across the four datasets, the range of $\kappa$ is 0.211 for LPSGM, 0.179 for SleepXViT and 0.127 for SleepVLM. The 0.020 average margin in $\kappa$ therefore comes from the absence of a weak dataset, not from a large lead on any single one.

We attribute this stability to rule grounding. It allows the model to cite only the features specified by the AASM rules, which limits its reliance on patterns specific to any single dataset. SleepVLM differs from the compared methods in that it is the only one that reports the AASM rules behind every staged epoch.

\subsection{Reasoning Quality}
\begin{figure}[t]
  \centering
  \includegraphics[width=\columnwidth]{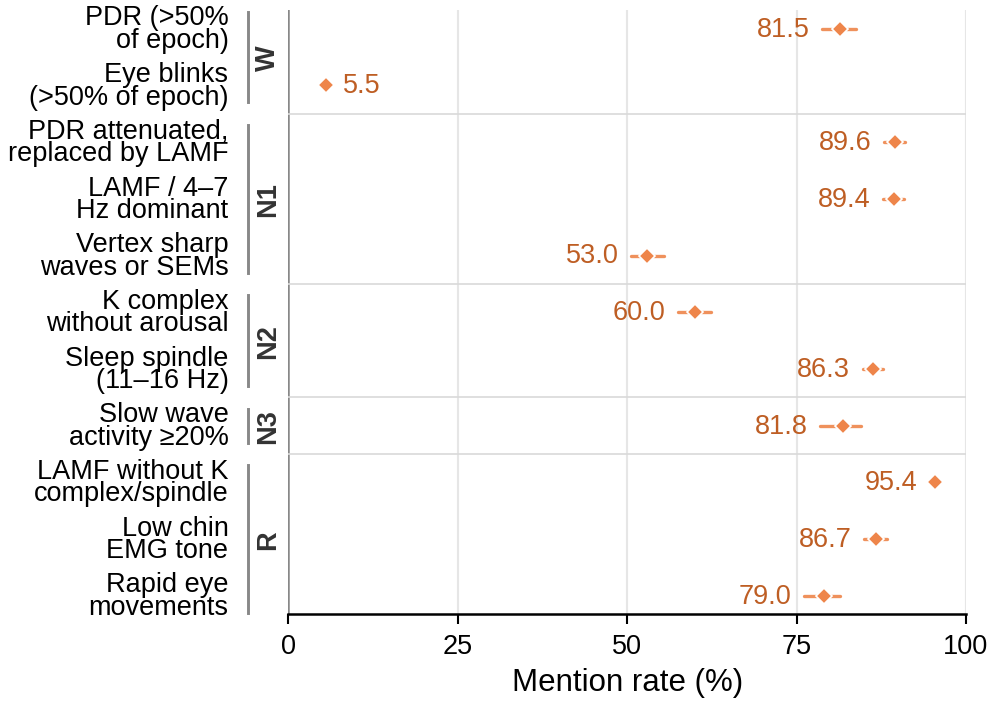}
  \caption{Coverage of AASM stage-defining features in SleepVLM rationales. Error bars are 95\% confidence intervals from 1,000 subject-level cluster bootstrap resamples.}
  \label{fig:audit}
\end{figure}
\begin{figure}[t]
  \centering
  \includegraphics[width=\columnwidth]{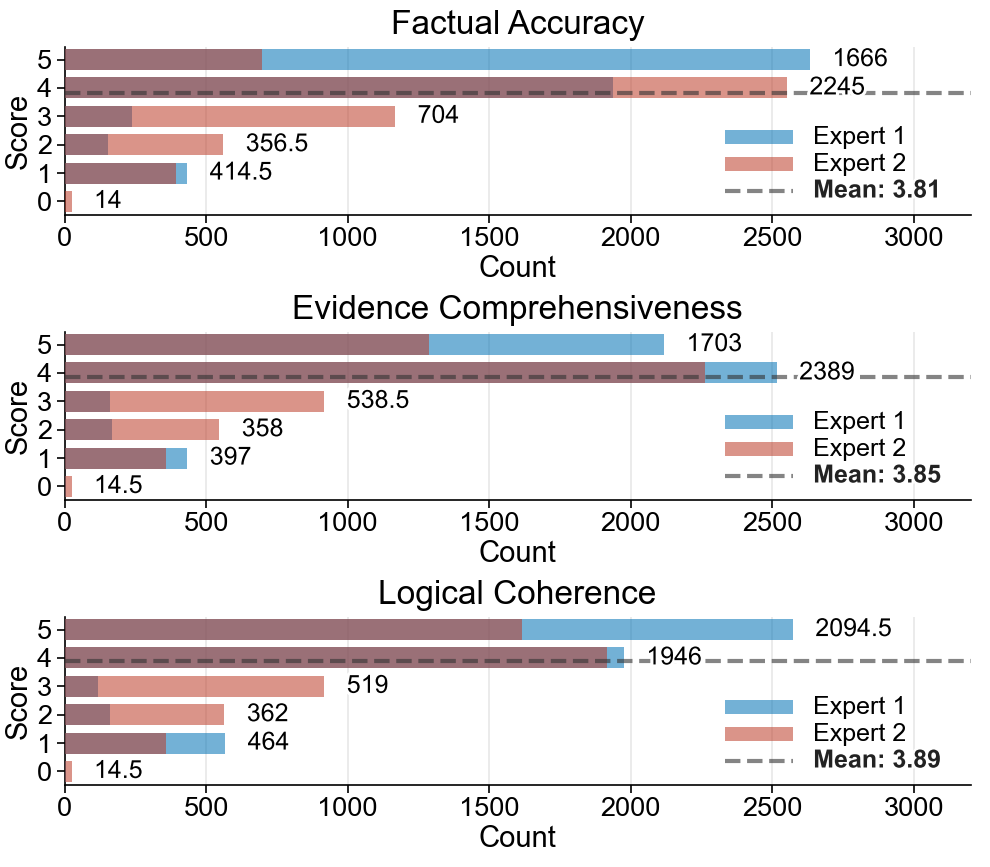}
  \caption{Expert ratings of SleepVLM reasoning quality. Labels are counts averaged over the two raters.}
  \label{fig:reason}
\end{figure}
\begin{table*}[t]
\centering\fontsize{7.8}{9.3}\selectfont
\setlength{\tabcolsep}{2pt}
\caption{Ablation study. \checkmark and $\times$ mark whether a component is used. Best value per column in bold.}
\label{tab:ablation}
\begin{tabular}{c cccc ccc ccc ccc ccc}
\toprule
\multirow{2}{*}[-\hdrvshift]{Configuration} & \multirow{2}{*}[-\hdrvshift]{WPT} & \multirow{2}{*}[-\hdrvshift]{SFT data} & \multirow{2}{*}[-\hdrvshift]{\shortstack{Rule\\grounding}} & \multirow{2}{*}[-\hdrvshift]{RFT} & \multicolumn{3}{c}{MASS-SS1} & \multicolumn{3}{c}{ZUAMHCS} & \multicolumn{3}{c}{ABC} & \multicolumn{3}{c}{DCSM} \\
\cmidrule(lr){6-8}\cmidrule(lr){9-11}\cmidrule(lr){12-14}\cmidrule(lr){15-17}
 & & & & & Acc & MF1 & $\kappa$ & Acc & MF1 & $\kappa$ & Acc & MF1 & $\kappa$ & Acc & MF1 & $\kappa$ \\
\midrule
Zero-shot baseline & $\times$ & -- & $\times$ & $\times$ & 0.100 & 0.070 & $-0.003$ & 0.201 & 0.104 & $-0.013$ & 0.093 & 0.120 & 0.008 & 0.092 & 0.119 & 0.008 \\
\textbf{SleepVLM} & \checkmark & Fine + Coarse & \checkmark & $\times$ & \textbf{0.835} & \textbf{0.793} & \textbf{0.767} & \textbf{0.812} & \textbf{0.766} & \textbf{0.743} & \textbf{0.739} & \textbf{0.706} & \textbf{0.640} & 0.789 & \textbf{0.733} & 0.717 \\
w/o WPT & $\times$ & Fine + Coarse & \checkmark & $\times$ & 0.818 & 0.753 & 0.736 & 0.804 & 0.743 & 0.727 & 0.687 & 0.656 & 0.575 & 0.758 & 0.686 & 0.677 \\
w/o Rule Grounding & \checkmark & Fine + Coarse & $\times$ & $\times$ & 0.829 & 0.785 & 0.758 & 0.795 & 0.746 & 0.720 & 0.704 & 0.651 & 0.581 & 0.770 & 0.700 & 0.686 \\
+RFT & \checkmark & Fine + Coarse & \checkmark & \checkmark & 0.810 & 0.738 & 0.728 & 0.801 & 0.737 & 0.727 & 0.681 & 0.605 & 0.547 & \textbf{0.795} & 0.712 & \textbf{0.718} \\
w/o Coarse Annotation & \checkmark & Fine only & \checkmark & $\times$ & 0.767 & 0.700 & 0.667 & 0.760 & 0.680 & 0.668 & 0.657 & 0.603 & 0.522 & 0.732 & 0.661 & 0.636 \\
w/o Coarse Annot.\ \& WPT & $\times$ & Fine only & \checkmark & $\times$ & 0.673 & 0.584 & 0.526 & 0.699 & 0.571 & 0.565 & 0.602 & 0.532 & 0.427 & 0.616 & 0.538 & 0.476 \\
\bottomrule
\end{tabular}
\end{table*}
\begin{figure*}[t]
  \centering
  \includegraphics[width=\textwidth]{figures/fig4_qualitative.png}
  \caption{Representative SleepVLM outputs.}
  \label{fig:qual}
\end{figure*}

A fluent rationale is not necessarily a grounded one. We assessed the reasoning quality of SleepVLM under two independent protocols: an automated AASM-feature audit and expert ratings.

\subsubsection{Automated audit}
We audited every rationale produced on the four evaluation datasets to check whether it describes the AASM stage-defining features for the ground-truth stage. The criteria are the 11 stage-defining features decomposed from the stage-onset rules; an LLM judge (Qwen3.6-plus) decided for each of them whether the rationale states it affirmatively, and the mention rate is the proportion of affirmative verdicts. Mention rates are at least 79\% for eight of the 11 features (Figure~\ref{fig:audit}). The highest is low-amplitude mixed-frequency (LAMF) EEG without K complexes or sleep spindles for stage R, at 95.4\%. Three features fall below this level: eye blinks at 5.5\%, vertex sharp waves or slow eye movements (SEMs) at 53.0\%, and K complexes unassociated with arousal at 60.0\%. All three are alternative onset cues for their stage rather than mandatory ones. A rationale already satisfies the criterion once it cites the more commonly used alternative: posterior dominant rhythm (PDR) for W, LAMF activity for N1, or a sleep spindle for N2. When epochs are stratified by classification correctness, mention rates for these features are substantially higher in correctly classified epochs than in incorrectly classified ones. The audit therefore shows not only that the stage-defining features are stated, but also that these statements track the model's own decisions rather than being stage-independent boilerplate. Further details are given in the Appendix.

\subsubsection{Expert ratings}
Two trained sleep technologists rated the reasoning quality of the rationales. For each of the 540 subjects, 10 epochs were drawn by stratified random sampling across sleep stages, which yielded 5,400 evaluation samples. Each technologist reviewed every sample independently against the rendered PSG image and the complete model output (the stage, the rule identifiers and the rationale), and scored factual accuracy, evidence comprehensiveness and logical coherence on an integer scale of 0 to 5. The rating scale is given in the Appendix. Averaged over the two raters and all samples, SleepVLM scores 3.81, 3.85 and 3.89 on the three dimensions (Figure~\ref{fig:reason}). These scores fall in the ``Acceptable'' to ``Good'' band of the scale. The distributions concentrate at the upper end: scores of 4--5 account for 72.4\%, 75.8\% and 74.8\% of the samples, and scores of 0--1 for 7.9\%, 7.6\% and 8.9\%. Agreement between the two raters is measured by a pooled ICC(2,k) of 0.663 (95\% CI 0.306--0.806) \cite{koo2016iccguideline,cicchetti1994guidelines}. The two protocols therefore point to the same conclusion: the rationales are grounded in the rules and in the images, not merely fluent. Representative outputs are shown in Figure~\ref{fig:qual}.

\subsection{Ablation Study}

To assess the contribution of each training component, we conducted an ablation study over seven configurations (Table~\ref{tab:ablation}). The zero-shot backbone reaches a $\kappa$ of only $-0.013$ to 0.008 across the four datasets, which is near chance and confirms that domain-specific fine-tuning is essential.

The three components contribute in different patterns. Removing WPT reduces $\kappa$ on every dataset, by 0.031, 0.016, 0.065 and 0.040 on MASS-SS1, ZUAMHCS, ABC and DCSM, respectively. Pre-training on per-second spectral and amplitude targets strengthens the visual feature extraction of the model. Removing rule grounding reduces $\kappa$ by only 0.009 on MASS-SS1, but by 0.023, 0.059 and 0.031 on the three external test sets; the loss is concentrated on the external sets, which indicates that explicit rule anchoring acts as a form of knowledge regularization whose benefit lies in cross-domain generalization. This is consistent with the stability observed in Table~\ref{tab:main}. Restricting SFT to the five subjects with fine-grained annotations reduces $\kappa$ by 0.100, 0.075, 0.118 and 0.081, and removing WPT as well leaves it at least 0.178 below SleepVLM on every dataset. Coarse annotations carry no rationale, but they extend supervision from 5 subjects to 50, and producing one rule identifier and one stage label costs far less than writing a rationale, which makes this the component that scales.

We also evaluated a third stage, rejection sampling fine-tuning (RFT). Starting from the best SFT checkpoint, it resamples rationales for the epochs with coarse annotations, keeps only the candidates whose predicted stage and rule set both match the annotation, and selects among them by perplexity gain. The procedure is given in the Appendix. +RFT brings no improvement over SleepVLM: $\kappa$ drops by 0.039, 0.016 and 0.093 on MASS-SS1, ZUAMHCS and ABC, and rises by 0.001 on DCSM. Rationales selected by perplexity gain may introduce distribution shift in this additional training stage. SleepVLM therefore keeps two stages. The three components it retains address different limitations: pre-training strengthens perception, rule grounding improves cross-domain generalization, and coarse annotations extend supervision coverage.

\subsection{Deployment Efficiency}
\begin{table}[t]
\centering\fontsize{7.8}{9.3}\selectfont
\setlength{\tabcolsep}{1.5pt}
\caption{Quantization comparison.}
\label{tab:quant}
\begin{tabular}{cc ccc @{\hspace{6pt}} ccc}
\toprule
\multicolumn{2}{c}{Base model} & \multicolumn{3}{c}{Qwen2.5-VL-3B} & \multicolumn{3}{c}{Qwen3.5-0.8B} \\
\cmidrule(lr){3-5}\cmidrule(lr){6-8}
\multicolumn{2}{c}{Precision} & BF16 & W4A16 & $\Delta$ & BF16 & W4A16 & $\Delta$ \\
\midrule
\multicolumn{2}{c}{Model size (GB)}          & 7.1 & 3.2 & -54.9\% & 1.7 & 1.2 & -29.4\% \\
\multicolumn{2}{c}{Inference speed (epoch/s)} & 9.4 & 9.9 & +5.3\% & 14.3 & 13.8 & -3.5\% \\
\midrule
\multirow{3}{*}{MASS-SS1} & Acc & 0.835 & 0.827 & -0.008 & 0.832 & 0.832 & 0.000 \\
                          & MF1 & 0.793 & 0.788 & -0.005 & 0.773 & 0.774 & +0.001 \\
                          & $\kappa$    & 0.767 & 0.757 & -0.010 & 0.761 & 0.760 & -0.001 \\
\midrule
\multirow{3}{*}{ZUAMHCS}  & Acc & 0.812 & 0.798 & -0.014 & 0.817 & 0.809 & -0.008 \\
                          & MF1 & 0.766 & 0.751 & -0.015 & 0.764 & 0.756 & -0.008 \\
                          & $\kappa$    & 0.743 & 0.727 & -0.016 & 0.749 & 0.739 & -0.010 \\
\midrule
\multirow{3}{*}{ABC}      & Acc & 0.739 & 0.728 & -0.011 & 0.730 & 0.722 & -0.008 \\
                          & MF1 & 0.706 & 0.697 & -0.009 & 0.693 & 0.696 & +0.003 \\
                          & $\kappa$    & 0.640 & 0.627 & -0.013 & 0.624 & 0.620 & -0.004 \\
\midrule
\multirow{3}{*}{DCSM}     & Acc & 0.789 & 0.776 & -0.013 & 0.775 & 0.765 & -0.010 \\
                          & MF1 & 0.733 & 0.712 & -0.021 & 0.716 & 0.706 & -0.010 \\
                          & $\kappa$    & 0.717 & 0.700 & -0.017 & 0.698 & 0.687 & -0.011 \\
\bottomrule
\end{tabular}
\end{table}

We reduce deployment cost along two directions: post-training quantization and a smaller backbone trained with the same two-stage pipeline. We quantize the linear layers of the language model to W4A16 (4-bit weights, 16-bit activations) with Intel AutoRound \cite{cheng2024autoround}, while the vision encoder and the language modeling head are kept in float16. Calibration uses 5,000 samples drawn from the training set with stratification across sleep stages. Both backbones, Qwen2.5-VL-3B and Qwen3.5-0.8B, were evaluated before and after quantization, with inference on a single NVIDIA RTX 5090 GPU (Table~\ref{tab:quant}). The full quantization setting is given in the Appendix.

Quantization shrinks the 3B model from 7.1\,GB to 3.2\,GB and the 0.8B model from 1.7\,GB to 1.2\,GB, at a cost of at most 0.017 and 0.011 in $\kappa$ across the four datasets, respectively. Switching to the 0.8B backbone at the same precision changes $\kappa$ by at most 0.019, and even raises it slightly on ZUAMHCS. Combining the two directions gives the smallest configuration: 1.2\,GB, 83\% smaller than the unquantized 3B model, with $\kappa$ at most 0.030 below it on the four datasets. Quantization changes throughput by at most 5.3\%, and moving from the 3B to the 0.8B backbone raises it from 9.4 to 14.3 epochs per second.

\section{Conclusion}

In this paper, we propose the task of auditable sleep staging and present SleepVLM, a vision-language model that casts sleep staging as visual reasoning over rendered PSG waveform images. For every epoch, it outputs a stage, the identifiers of the applicable AASM rules, and an auditable rationale describing the stage-defining features it observes. We obtain this behavior in two training stages. Waveform-perceptual pre-training teaches the model to interpret waveform images quantitatively. Rule-grounded supervised fine-tuning then writes all 15 AASM rules into the system prompt and requires the model to cite the identifiers that support each decision. To supply this supervision, we construct and release MASS-EX, to our knowledge the first expert-annotated dataset for rule-grounded sleep staging, which annotates 62 subjects with the applicable rule identifiers and adds expert-written rationales for a subset. On four datasets, SleepVLM attains the highest average accuracy, macro-F1 and $\kappa$, with no weak dataset among them. An automated AASM-feature audit and independent ratings by two trained sleep technologists over 5,400 samples show that its rationales are grounded in the rules and in the images, not merely fluent. Unlike post-hoc explanation methods, which leave a scorer to reinterpret where a model attends, SleepVLM reports the rules it applies together with the stage, so its rationale can be audited directly against the AASM manual.

\bibliography{references}

\end{document}


\setcounter{page}{11}
\maketitle

\noindent
This document supplements the main paper. Seven of its sections answer the
places where the main paper refers to the Appendix.

\begin{table}[h]
\centering
\small
\caption{Where each reference in the main paper is answered.}
\label{tab:map}
\begin{tabular}{@{}p{0.26\linewidth} p{0.46\linewidth} l@{}}
\toprule
Main paper section & Reference & Here \\
\midrule
MASS-EX Dataset & ``The complete rule set is given in the Appendix.'' & \S\ref{sec:rules} \\
MASS-EX Dataset & ``The full pipeline is given in the Appendix.'' & \S\ref{sec:pipeline} \\
Method & ``The complete prompt and both output templates are given in the Appendix.'' & \S\ref{sec:prompts} \\
Experiments & ``Further details are given in the Appendix.'' & \S\ref{sec:audit} \\
Experiments & ``The rating scale is given in the Appendix.'' & \S\ref{sec:ratings} \\
Experiments & ``The procedure is given in the Appendix.'' & \S\ref{sec:rft} \\
Experiments & ``The full quantization setting is given in the Appendix.'' & \S\ref{sec:hparams} \\
\bottomrule
\end{tabular}
\end{table}

\noindent
The remaining sections report information that did not fit within the page
limit of the main paper: the datasets (\S\ref{sec:datasets}), the evaluation
metrics (\S\ref{sec:metrics}), the experimental setup (\S\ref{sec:setup}),
detailed results with confidence intervals (\S\ref{sec:perclass}) and further
qualitative examples (\S\ref{sec:qual}).

\clearpage

\section{Rule Set}
\label{sec:rules}

\noindent
The complete rule set referred to in the main paper is given in
Table~\ref{tab:rules}.

{\footnotesize
\setlength{\tabcolsep}{4pt}
\begin{longtable}{|L{0.105\textwidth}|L{0.068\textwidth}|L{0.125\textwidth}|L{0.512\textwidth}|L{0.100\textwidth}|}
\caption{AASM sleep staging rules operationalized for visual reasoning.}
\label{tab:rules}\\
\hline
\textbf{Stage Category} & \textbf{Rule ID} & \textbf{Rule Type} &
\textbf{Operationalized Visual Criteria} & \textbf{Assigned Stage} \\
\hline
\endfirsthead
\multicolumn{5}{l}{\footnotesize\itshape Table \thetable\ (continued)}\\
\hline
\textbf{Stage Category} & \textbf{Rule ID} & \textbf{Rule Type} &
\textbf{Operationalized Visual Criteria} & \textbf{Assigned Stage} \\
\hline
\endhead
\hline
\endfoot
\hline
\endlastfoot
\multirow{3}{*}{W} & W.1 & Onset & $>$50\% of the epoch contains posterior
dominant rhythm (alpha rhythm, 8--13 Hz), primarily measured in the occipital
derivation (O2-M1). & W \\
\cline{2-5}
 & W.2 & Onset & $>$50\% of the epoch contains eye blinks (conjugate vertical
movements at 0.5--2 Hz) in EOG channels. & W \\
\cline{2-5}
 & W.3 & Onset & $>$50\% of the epoch contains irregular, conjugate rapid eye
movements (REMs) in EOG channels associated with normal or high chin muscle
tone. & W \\
\hline
\multirow{2}{*}{N1} & N1.1 & Onset & For alpha generators: Posterior dominant
rhythm is attenuated and replaced by LAMF activity for $>$50\% of the epoch.
& N1 \\
\cline{2-5}
 & N1.2 & Onset & For non-alpha generators: Earliest appearance of EEG slowing
by $\ge$1 Hz from stage W (into 4--7 Hz range), vertex sharp waves, or slow eye
movements (SEMs). & N1 \\
\hline
\multirow{4}{*}{N2} & N2.1 & Onset & In the absence of criteria for stage N3,
presence of $\ge$1 K complexes unassociated with arousals or $\ge$1 sleep
spindles (11--16 Hz) occurring in the first half of the epoch or the last half
of the preceding epoch. & N2 \\
\cline{2-5}
 & N2.2 & Continuation & Epoch exhibits LAMF activity without K complexes or
sleep spindles, but is preceded by an epoch containing non-arousal associated
K complexes or sleep spindles without an intervening arousal. & N2 \\
\cline{2-5}
 & N2.3 & Continuation & Epoch following an N3 stage that no longer meets the
criteria for stage N3, lacks an intervening arousal, and does not meet criteria
for stage W or stage R. & N2 \\
\cline{2-5}
 & N2.4 & Termination & End N2 when transitioning to stage W, stage N3, or
stage R; upon an arousal followed by LAMF (revert to N1 until a K complex
unassociated with an arousal or a sleep spindle reappears); or upon a major
body movement followed by SEMs (revert to N1). & W, N1, N3, or R \\
\hline
N3 & N3.1 & Onset & $\ge$20\% of the epoch consists of slow wave activity
(0.5--2.0 Hz, peak-to-peak amplitude $>$75\,\textmu{}V), predominantly in the
frontal derivations (F4-M1). & N3 \\
\hline
\multirow{3}{*}{R} & R.1 & Onset & Definite R: Simultaneous presence of
(a)~LAMF EEG without K complexes or sleep spindles, (b)~low chin EMG tone for
the majority of the epoch and concurrent with REMs, and (c)~REMs at any
position within the epoch. & R \\
\cline{2-5}
 & R.2 & Continuation & Epoch contiguous with a definite R epoch showing
persistent LAMF EEG without K complexes or sleep spindles and low chin EMG
tone, with no intervening arousals and in the absence of REMs or SEMs following
an arousal. & R \\
\cline{2-5}
 & R.3 & Termination & End R when transitioning to W or N3; chin EMG tone
increases above the level of stage R for the majority of the epoch with N1-like
EEG; an arousal occurs followed by SEMs (score as N1 even if chin EMG remains
low); or K complexes/sleep spindles appear in the first half of the epoch
without REMs (score as N2). & W, N1, N2, or N3 \\
\hline
\multirow{2}{\linewidth}{Major Body Movement} & MBM.1 &
Scoring & Epoch is obscured by movement and muscle artifact for more than half
an epoch. Scored as W if posterior dominant rhythm (alpha rhythm) is present
for part of the epoch, or if an epoch scoreable as stage W either precedes or
follows the epoch. & W \\
\cline{2-5}
 & MBM.2 & Scoring & Epoch obscured by artifact not meeting MBM.1 criteria.
Scored as the same stage as the epoch that follows it. & Same as subsequent \\
\end{longtable}
}

\vspace{-0.6em}
\noindent
{\footnotesize
Fifteen rules for adult sleep staging adapted from the AASM Manual for the
Scoring of Sleep and Associated Events (Version 3) \citep{berry2023aasm},
covering the rules
applicable to the three EEG (F4-M1, C4-M1, O2-M1), two EOG (LOC, ROC), and chin
EMG channels used in this study. For presentation, the rules shown here are
compressed versions rather than the full verbatim text. AASM American Academy
of Sleep Medicine, MBM major body movement, EEG electroencephalography, EOG
electrooculography, EMG electromyography, LAMF low-amplitude mixed-frequency,
SEMs slow eye movements, REMs rapid eye movements.
}

\section{Annotation Pipeline}
\label{sec:pipeline}

\noindent
MASS-EX keeps the sleep stage labels of MASS-SS3. The pipeline produced the
applicable rule identifiers of each epoch and the rationale of the fine-grained
annotations. The fine-grained annotations come from the first five subjects,
01-03-0001 to 01-03-0005, and cover 5,006 epochs; the other 57 subjects and
their 54,187 epochs received rule identifiers only. The annotations were
produced by a trained sleep technologist and a senior sleep medicine physician
with over a decade of clinical experience. The 15 rules of
Section~\ref{sec:rules} were developed jointly by the same two experts.

\paragraph{Exemplar annotations.}
The two experts first selected 16 epochs, so that their applicable rules
together cover all 15 rules of the rule set. For each of these epochs they wrote
an exemplar annotation, including the applicable rule identifiers and the
rationale. The exemplar rationales follow a fixed order: they first describe the waveforms
observed in each channel, then point out the features that support the staging
decision, then cite the corresponding rules, and finally give the staging
conclusion. These 16 exemplars were then used as the few-shot demonstrations for
draft generation.

\paragraph{Draft generation.}
A locally deployed Qwen2.5-VL-72B-Instruct model \citep{bai2025qwen25vl}, with
these 16 exemplars as few-shot demonstrations, generated one draft rationale for
each of the 5,006 epochs. The drafts did not enter the dataset; their purpose
was to give the technologist a text to revise rather than write each rationale
from the start.

\paragraph{Technologist review.}
The sleep technologist then went through the 5,006 drafts one by one. The
waveform image of the epoch was the reference: every statement in a draft about
a channel, timing, morphology, frequency or amplitude was compared with the
image, and corrected where the two did not agree. The cited rules were checked
in the same way, and the applicable rule identifiers were decided by the
technologist. Of the 5,006 drafts, 4,639 (92.7\%) were edited. The remaining
54,187 epochs needed only the applicable rule identifiers, and the technologist
annotated them directly.

\paragraph{Physician verification.}
The senior physician then reviewed all annotations of both subsets one by one
and directly corrected the entries judged incorrect. The final annotations
follow the physician's corrections.

\section{Prompts}
\label{sec:prompts}

Two system prompts were used, one for each training stage. Throughout this
section, shaded boxes contain the verbatim prompt text provided to the model;
unshaded text provides editorial context for the reader.

\subsection{Waveform-Perceptual Pre-training}

The model received a single 30-second PSG waveform image and was trained to
predict per-second spectral and amplitude features for each visible channel.

\begin{promptbox}
\textbf{\# Task}

Analyze a 30-second PSG waveform image and estimate the following features at
each second:
\begin{ptlist}
\item \textbf{EEG/EOG channels:} frequency band power (delta, theta, alpha,
      beta) in dB, plus signal amplitude (MAV) in \textmu{}V
\item \textbf{EMG channel (Chin):} muscle tone (MAV) in \textmu{}V
\end{ptlist}

\textbf{\# Image Rendering Parameters}

You must interpret the image according to the following fixed amplitude
scales:
\begin{ptlist}
\item \textbf{EEG (F4-M1, C4-M1, O2-M1):} $-$50 \textmu{}V to $+$50 \textmu{}V
\item \textbf{EOG (LOC, ROC):} $-$50 \textmu{}V to $+$50 \textmu{}V
\item \textbf{Chin EMG:} $-$40 \textmu{}V to $+$40 \textmu{}V
\end{ptlist}

\textbf{Channel-to-color mapping:} F4-M1 (yellow), C4-M1 (green), O2-M1 (red),
LOC (cyan), ROC (magenta), Chin EMG (blue).

\textbf{Channel order:} from top to bottom: F4-M1, C4-M1, O2-M1, LOC, ROC,
Chin EMG.

\textbf{Note:} Some channels may be missing in the image. Only output channels
that are visible in the image.

\textbf{\# Output Format}
\begin{verbatim}
{
  "F4-M1": [[d,t,a,b,mav], ...],    // 30 arrays, 5 values each
  "C4-M1": [[d,t,a,b,mav], ...],
  "O2-M1": [[d,t,a,b,mav], ...],
  "LOC":   [[d,t,a,b,mav], ...],
  "ROC":   [[d,t,a,b,mav], ...],
  "Chin":  [[mav], ...]              // 30 arrays, 1 value each
}
\end{verbatim}

\textbf{EEG/EOG channels:} Each contains 30 arrays (seconds 1--30), with 5
values per array: [delta, theta, alpha, beta, mav]. The first 4 values are
band powers in dB; the 5th value (mav) is the Mean Absolute Value of signal
amplitude in \textmu{}V. All values use 1 decimal place.

\textbf{Chin EMG channel:} Contains 30 arrays (seconds 1--30), with 1 value
per array: [mav] representing muscle tone via Mean Absolute Value in
\textmu{}V (1 decimal place).
\end{promptbox}

\subsection{Rule-Grounded Supervised Fine-tuning}

The fine-grained and coarse annotation training tracks shared an identical
prompt, with the exception of Sections 1, 4 and 5, noted in-line below.

\subsubsection*{Section 1: Roles and Tasks}

The prompt box below is the fine-grained version. The coarse annotation track
omits ``detailed, precise'' before ``rule-based rationale''.

\begin{promptbox}
You are an experienced sleep technician. Your task is to analyze an image
sequence containing three consecutive 30-second PSG epochs and provide an
accurate sleep stage and a detailed, precise, rule-based rationale for the
central target epoch.
\end{promptbox}

\subsubsection*{Section 2A: Image Rendering Parameters}

\begin{promptbox}
You must interpret the images according to the following fixed amplitude
scales:
\begin{ptlist}
\item \textbf{EEG (F4-M1, C4-M1, O2-M1):} $-$50 \textmu{}V to $+$50 \textmu{}V
\item \textbf{EOG (LOC, ROC):} $-$50 \textmu{}V to $+$50 \textmu{}V
\item \textbf{Chin EMG:} $-$40 \textmu{}V to $+$40 \textmu{}V
\end{ptlist}

\textbf{Channel-to-color mapping:} F4-M1 (yellow), C4-M1 (green), O2-M1 (red),
LOC (cyan), ROC (magenta), Chin EMG (blue).

\textbf{Channel order:} from top to bottom: F4-M1, C4-M1, O2-M1, LOC, ROC,
Chin EMG.

\textbf{Key interpretation notes:}

\textbf{High-amplitude estimation:} If the vertical amplitude of a slow wave
occupies more than 75\% of its channel height, it meets the amplitude
criterion for stage N3 ($>$75 \textmu{}V peak-to-peak).

\textbf{Signal overlap as corroboration:} A clear visual indication of
extremely high amplitude is when a waveform physically overlaps or extends
into the display area of an adjacent channel. You must interpret such overlap
as definitive evidence that the amplitude exceeds the channel boundary and
satisfies the $>$75 \textmu{}V threshold.
\end{promptbox}

\subsubsection*{Section 2B: AASM Version 3 Sleep Staging Rules}

The system prompt included the complete text of all 15 AASM Version 3 adult
sleep staging rules, provided verbatim to the model. This section comprised
approximately 1,000 words and covered: general scoring principles (epoch
scoring, dominance principle, primary channels); definitions of key waveforms
and events (alpha rhythm, theta waves, slow wave activity, eye blinks, REMs,
SEMs, K complexes, sleep spindles, low chin EMG tone, sawtooth waves,
arousals); and the full scoring criteria for each stage (Rules W.1--W.3,
N1.1--N1.2, N2.1--N2.4, N3.1, R.1--R.3, MBM.1--MBM.2). To avoid redundancy,
the rule text is not reproduced here; the operationalized version of these
rules is given in Section~\ref{sec:rules}.

\subsubsection*{Section 3: Input Data}

\begin{promptbox}
Below, an image sequence containing three consecutive 30-second PSG epochs
will be provided, namely:
\begin{ptlist}
\item the preceding epoch N-1
\item the target epoch N (the central image)
\item the subsequent epoch N+1
\end{ptlist}

Your analysis must focus on the central image labeled as the target epoch N,
but you may consider the preceding and subsequent epochs to understand
dynamic changes.
\end{promptbox}

\subsubsection*{Section 4: Tasks and Instructions}

The fine-grained annotation track includes all five steps below. The coarse
annotation track omits Step 4; this difference is marked with a bracket
annotation in the prompt box.

\begin{promptbox}
Please strictly follow the steps below:
\begin{ptenum}
\item \textbf{Analyze the context:} Using the preceding (epoch N-1) and
      subsequent (epoch N+1) epoch images, analyze the dynamics of the target
      epoch N. Note that this is to understand trends, but your final
      judgement must be based on observing the target epoch N itself.
\item \textbf{Identify key features:} In the target epoch N, identify all key
      waveforms and features in great detail. Your description must include:
      channel names, occurrence times, waveform types, frequencies, amplitude
      estimates.
\item \textbf{Cite rules:} Explicitly identify the specific AASM rule numbers
      from the knowledge base that support your classification.
\item \textbf{Generate the rationale} [fine-grained track only]: Integrate
      your findings into a professional, concise rationale written without
      first-person pronouns. You must include quantitative or
      semi-quantitative descriptions of amplitude, frequency, duration, etc.
\item \textbf{Format the output:} Your final analysis result may only be
      provided in the following JSON format.
\end{ptenum}
\end{promptbox}

\subsubsection*{Section 5: Output Format}

This is the only section where the two training tracks diverge in their
expected output schema. Both JSON templates are shown below.

\noindent
(a) Fine-grained annotation track
\begin{shaded}
\begin{verbatim}
{
  "reasoning_text": "<Provide an extremely detailed description of the
    key waveform features in the EEG, EOG and EMG channels of the target
    epoch N, including but not limited to: channel names, occurrence times,
    waveform types, frequencies, amplitude estimates, and relationships
    with adjacent channels (e.g., overlap). Then, based on these
    observations, apply the AASM Version 3 rules to logically infer why
    the epoch belongs to the specified stage. The text must be
    professional, objective and coherent.>",
  "applicable_rules": ["<AASM rule numbers, e.g., W.1, N2.1, R.2>"],
  "sleep_stage": "<W/N1/N2/N3/R>"
}
\end{verbatim}
\end{shaded}

\noindent
(b) Coarse annotation track
\begin{shaded}
\begin{verbatim}
{
  "applicable_rules": ["<AASM rule numbers, e.g., W.1, N2.1, R.2>"],
  "sleep_stage": "<W/N1/N2/N3/R>"
}
\end{verbatim}
\end{shaded}

\noindent
The sole difference between the two tracks is that the fine-grained track
includes the \mbox{\texttt{reasoning\_text}} field containing the model's complete
rationale, whereas the coarse track omits this field and requires only the
applicable rule identifiers and the predicted sleep stage. During inference,
the model always uses the fine-grained output schema to generate a full
rationale regardless of how it was trained.

\section{Hyperparameters}
\label{sec:hparams}

\noindent
The complete training, quantization and inference hyperparameters are given in
Table~\ref{tab:hparams}.

{\small
\setlength{\tabcolsep}{4pt}
\begin{longtable}{|L{0.30\textwidth}|L{0.209\textwidth}|L{0.209\textwidth}|L{0.209\textwidth}|}
\caption{Training, quantization and inference hyperparameters.}
\label{tab:hparams}\\
\hline
\textbf{Training} & \textbf{Phase 1 (WPT)} & \textbf{Phase 2 (SFT)} &
\textbf{Phase 3 (RFT)} \\
\hline
\endfirsthead
\multicolumn{4}{l}{\small\itshape Table \thetable\ (continued)}\\
\hline
\endhead
LoRA rank ($r$) / $\alpha$ / dropout & 16 / 32 / 0.05 & 16 / 32 / 0.05 &
16 / 32 / 0.05 \\
\hline
LoRA applied layers & All nn.Linear excluding lm\_head & nn.Linear in language
model, excluding lm\_head & nn.Linear in language model, excluding lm\_head \\
\hline
Vision encoder & Unfrozen & Frozen & Frozen \\
\hline
Optimizer & AdamW ($\beta_1$=0.9, $\beta_2$=0.95, $\epsilon$=$10^{-8}$) &
AdamW ($\beta_1$=0.9, $\beta_2$=0.95, $\epsilon$=$10^{-8}$) &
AdamW ($\beta_1$=0.9, $\beta_2$=0.95, $\epsilon$=$10^{-8}$) \\
\hline
Learning rate & $10^{-4}$ & $10^{-4}$ & $10^{-4}$ \\
\hline
Learning rate schedule & Linear warmup (3\%), linear decay & Linear warmup
(3\%), linear decay & Linear warmup (3\%), linear decay \\
\hline
Weight decay & 0.1 & 0.1 & 0.1 \\
\hline
Gradient clipping (max norm) & 1.0 & 1.0 & 1.0 \\
\hline
Epochs & 2 & 15 & 5 \\
\hline
Per-device batch size & 4 & 6 & 6 \\
\hline
Gradient accumulation steps & 1 & 1 & 1 \\
\hline
Effective batch size & 32 & 48 & 48 \\
\hline
\pagebreak
\multicolumn{4}{|l|}{\textbf{Post-training quantization}} \\
\hline
Framework & \multicolumn{3}{L{\dimexpr 0.627\textwidth+16.8pt\relax}|}{Intel
AutoRound (0.13.1)} \\
\hline
Scheme & \multicolumn{3}{L{\dimexpr 0.627\textwidth+16.8pt\relax}|}{W4A16
(4-bit integer weights, 16-bit floating-point activations)} \\
\hline
Quantized layers & \multicolumn{3}{L{\dimexpr 0.627\textwidth+16.8pt\relax}|}{Qwen2.5-VL-3B:
all nn.Linear in the 36 language model transformer blocks. Qwen3.5-0.8B: the
query, key, value and output projections of its 6 full-attention blocks, and the
gate, up and down projections of all 24 MLP blocks (96 of its 237 nn.Linear
modules)} \\
\hline
Layers retained at original precision &
\multicolumn{3}{L{\dimexpr 0.627\textwidth+16.8pt\relax}|}{Vision encoder,
lm\_head; for Qwen3.5-0.8B also its 18 gated delta linear-attention blocks} \\
\hline
Group size & \multicolumn{3}{L{\dimexpr 0.627\textwidth+16.8pt\relax}|}{128} \\
\hline
Optimization iterations &
\multicolumn{3}{L{\dimexpr 0.627\textwidth+16.8pt\relax}|}{200} \\
\hline
Calibration samples & \multicolumn{3}{L{\dimexpr 0.627\textwidth+16.8pt\relax}|}{5,000
(stratified by sleep stage from SFT training set)} \\
\hline
Calibration sequence length &
\multicolumn{3}{L{\dimexpr 0.627\textwidth+16.8pt\relax}|}{3,140 tokens} \\
\hline
\multicolumn{4}{|l|}{\textbf{Inference}} \\
\hline
Framework & \multicolumn{3}{L{\dimexpr 0.627\textwidth+16.8pt\relax}|}{vLLM} \\
\hline
Precision & \multicolumn{3}{L{\dimexpr 0.627\textwidth+16.8pt\relax}|}{bfloat16
(float16 for quantized model)} \\
\hline
Temperature &
\multicolumn{3}{L{\dimexpr 0.627\textwidth+16.8pt\relax}|}{$10^{-6}$} \\
\hline
Top-p & \multicolumn{3}{L{\dimexpr 0.627\textwidth+16.8pt\relax}|}{0.8} \\
\hline
Maximum new tokens &
\multicolumn{3}{L{\dimexpr 0.627\textwidth+16.8pt\relax}|}{1024} \\
\hline
\end{longtable}
}

\vspace{-0.6em}
\noindent
{\footnotesize
Training was performed on a single node with 8 NVIDIA A100 GPUs (80\,GB each)
using bfloat16 mixed precision. Phase 3 (RFT) is the third stage evaluated as an
ablation in the main paper. The smaller backbone in the main paper, Qwen3.5-0.8B, uses the same
training and inference settings; only its quantized layers differ, because 18 of
its 24 blocks use gated delta linear attention instead of full attention. WPT,
waveform-perceptual pre-training; SFT, rule-grounded supervised fine-tuning;
RFT, rejection sampling fine-tuning.
\par}

\section{Rejection Sampling Fine-tuning}
\label{sec:rft}

\noindent
Rejection sampling fine-tuning (RFT) is designed to expand reasoning
supervision for the subjects with coarse annotations without requiring
additional expert annotation. Starting from the best SFT checkpoint, we perform
60 independent stochastic inference runs for every epoch from the 45 subjects
with coarse annotations, using temperature = 1.0 and top-$p$ = 1.0. A candidate
response is retained only if three conditions are satisfied: (1) the response
is successfully parsed into the predicted sleep stage, explanatory reasoning,
and applicable rule identifiers; (2) the predicted sleep stage exactly matches
the ground-truth label and the predicted rule set exactly matches the
ground-truth rule set; and (3) the rationale is written in English only.

Among the retained correct candidates for each epoch, the best rationale is
selected using a perplexity-gain criterion. For a candidate reasoning sequence
$r = (r_1, \dots, r_T)$ conditioned on context $c$, perplexity is defined as
\begin{equation}
\mathrm{PPL}(r \mid c) = \exp\!\left(-\frac{1}{T}\sum_{t=1}^{T}
  \log p(r_t \mid r_{<t}, c)\right)
\end{equation}
Perplexity-gain is then defined as
\begin{equation}
\text{PPL-gain} = \mathrm{PPL}(r \mid c_{\mathrm{full}})
  - \mathrm{PPL}(r \mid c_{\text{text-only}})
\end{equation}
where $c_{\mathrm{full}}$ denotes the full conditioning context including the
system prompt and the PSG images, and $c_{\text{text-only}}$ denotes the
same context with the PSG images removed. A lower PPL-gain indicates that the
rationale is more strongly conditioned on the visual input rather than
predictable from language priors alone. For each epoch, the correct response
with the minimum PPL-gain is selected as the training target. The resulting
perplexity-gain-selected reasoning samples are mixed with the original
fine-grained annotations from the five subjects that carry them to form the RFT
training set. Training uses the same optimization settings as SFT and runs for
5 training epochs; these are listed in Table~\ref{tab:hparams}.

\section{Datasets}
\label{sec:datasets}

\noindent
The characteristics of all datasets used in this work are given in
Table~\ref{tab:dschar} and their sleep stage distributions in
Table~\ref{tab:dsdist}. Each dataset is described below.

{\footnotesize
\setlength{\tabcolsep}{2pt}
\begin{longtable}{|L{0.100\textwidth}|L{0.030\textwidth}|L{0.099\textwidth}|L{0.064\textwidth}|L{0.073\textwidth}|L{0.119\textwidth}|L{0.111\textwidth}|L{0.073\textwidth}|L{0.064\textwidth}|L{0.172\textwidth}|}
\caption{Characteristics of datasets used in this study.}
\label{tab:dschar}\\
\hline
\textbf{Dataset} & \textbf{n} & \textbf{Age (mean $\pm$ SD)} &
\textbf{Age range} & \textbf{Sex (M / F)} &
\textbf{EEG/EOG/\allowbreak EMG elec\-trodes} & \textbf{Fs (Hz)} &
\textbf{Scoring stan\-dard} & \textbf{Epoch length (s)} &
\textbf{Use in this study} \\
\hline
\endfirsthead
\multicolumn{10}{l}{\footnotesize\itshape Table \thetable\ (continued)}\\
\hline
\textbf{Dataset} & \textbf{n} & \textbf{Age (mean $\pm$ SD)} &
\textbf{Age range} & \textbf{Sex (M / F)} &
\textbf{EEG/EOG/\allowbreak EMG elec\-trodes} & \textbf{Fs (Hz)} &
\textbf{Scoring stan\-dard} & \textbf{Epoch length (s)} &
\textbf{Use in this study} \\
\hline
\endhead
MASS-SS1 & 53 & 63.0 $\pm$ 5.3 & 55--76 & 34 / 19 & 17--19/2/3 &
256/256/256 & AASM & 30 & Held-out test set \\
\hline
MASS-SS2 & 19 & 23.6 $\pm$ 3.7 & 18--33 & 8 / 11 & 19/4/1 & 256/256/256 &
R\&K & 20 & WPT \\
\hline
MASS-SS3 & 62 & 42.5 $\pm$ 18.9 & 20--69 & 29 / 33 & 20/2/3 & 256/256/128 &
AASM & 30 & SFT and validation \\
\hline
MASS-SS4 & 40 & 25.3 $\pm$ 4.3 & 18--35 & 14 / 26 & 4/4/1 & 256/256/256 &
R\&K & 20 & WPT \\
\hline
MASS-SS5 & 26 & 25.0 $\pm$ 7.4 & 20--59 & 13 / 13 & 20/2/3 & 256/256/128 &
R\&K & 20 & WPT \\
\hline
ZUAMHCS & 100 & 28.4 $\pm$ 10.5 & 18--70 & 42 / 58 & 6/2/2--3 &
512/512/512 & AASM & 30 & External test set \\
\hline
ABC & 132 & 48.8 $\pm$ 9.9 & 27--65 & N/A & 6/2/3 & 256/256/256 &
AASM & 30 & External test set \\
\hline
DCSM & 255 & N/A & N/A & N/A & 6/2/1 & 256/256/256 & AASM & 30 &
External test set \\
\hline
\end{longtable}
}

\vspace{-0.6em}
\noindent
{\footnotesize
MASS, Montreal Archive of Sleep Studies. Fs, sampling frequency of EEG, EOG,
and EMG channels. R\&K, Rechtschaffen and Kales; AASM, American Academy of Sleep
Medicine.
\par}

{\footnotesize
\setlength{\tabcolsep}{3pt}
\begin{longtable}{|L{0.115\textwidth}|L{0.090\textwidth}|L{0.138\textwidth}|L{0.138\textwidth}|L{0.138\textwidth}|L{0.138\textwidth}|L{0.138\textwidth}|}
\caption{Sleep stage distribution across datasets used in this study.}
\label{tab:dsdist}\\
\hline
\textbf{Dataset} & \textbf{Total epochs} &
\textbf{W} & \textbf{N1} & \textbf{N2} & \textbf{N3} & \textbf{R} \\
\hline
\endfirsthead
\multicolumn{7}{l}{\footnotesize\itshape Table \thetable\ (continued)}\\
\hline
\textbf{Dataset} & \textbf{Total epochs} &
\textbf{W} & \textbf{N1} & \textbf{N2} & \textbf{N3} & \textbf{R} \\
\hline
\endhead
MASS-SS1 & 51,165 & 12,152 (23.8\%) & 7,098 (13.9\%) &
22,152 (43.3\%) & 3,402 (6.6\%) & 6,361 (12.4\%) \\
\hline
MASS-SS3 & 59,317 & 6,442 (10.9\%) & 4,839 (8.2\%) &
29,802 (50.2\%) & 7,653 (12.9\%) & 10,581 (17.8\%) \\
\hline
ZUAMHCS & 96,317 & 13,765 (14.3\%) & 7,059 (7.3\%) &
39,809 (41.3\%) & 19,549 (20.3\%) & 16,135 (16.8\%) \\
\hline
ABC & 133,000 & 30,938 (23.3\%) & 19,296 (14.5\%) &
52,334 (39.3\%) & 11,761 (8.8\%) & 18,671 (14.0\%) \\
\hline
DCSM & 304,266 & 79,636 (26.2\%) & 21,140 (6.9\%) &
113,027 (37.1\%) & 43,637 (14.3\%) & 46,826 (15.4\%) \\
\hline
\end{longtable}
}

\paragraph{MASS.}
MASS, the Montreal Archive of Sleep Studies \citep{oreilly2014mass}, is an
open-access database released by the Center for Advanced Research in Sleep Medicine in Montreal. It pools 200
whole-night laboratory polysomnograms from eight research protocols conducted in
three hospital sleep laboratories, and is organised into subsets. This work used all five of them: SS2, SS4 and SS5 for waveform-perceptual
pre-training, SS3 as the recordings that MASS-EX annotates and therefore as the
source of both the fine-tuning and the validation set, and SS1 as the held-out
test set.

\paragraph{ABC.}
ABC, the Apnea, Bariatric surgery, and CPAP study \citep{bakker2018gastric},
is distributed by the National Sleep Research Resource
\citep{zhang2018national}. It comes from a randomised trial that compared
laparoscopic gastric banding against continuous positive airway pressure in
patients with severe obstructive sleep apnea and class II obesity, and contains
132 overnight recordings collected at a baseline visit and at 9-month and
18-month follow-up visits. This work used it for external evaluation only.

\paragraph{DCSM.}
DCSM, the Danish Center for Sleep Medicine dataset, was released alongside the
U-Sleep study \citep{perslev2021uSleep}. It contains 255 randomly selected, fully de-identified overnight
laboratory polysomnograms recorded between 2015 and 2018 from patients referred
to that centre for the diagnosis of non-specific sleep-related disorders. EEG
and EOG were sampled at 256\,Hz and the hypnograms were scored according to AASM
criteria. Because the recordings are fully de-identified, no age or sex
information is distributed with them. This work used it for external evaluation
only.

\paragraph{ZUAMHCS.}
ZUAMHCS is a private clinical dataset comprising 100 adult subjects randomly
selected from patients who underwent clinical PSG between 2020 and 2025.
Recordings were acquired using a Compumedics Grael polysomnography system at a
sampling frequency of 512\,Hz and scored according to AASM criteria. ZUAMHCS
was used exclusively for external evaluation and was not involved in any
training, validation, checkpoint selection, or model design decisions. Data
collection was approved by an institutional review board, and written consent
was obtained from all subjects or their caregivers.

Most publicly available sleep datasets were collected for specific research
protocols and recorded under controlled laboratory conditions, so their patient
populations and acquisition settings differ from those of routine clinical
practice.
ZUAMHCS therefore complements the public evaluation sets rather than replacing
any of them, and extends the generalisation test to recordings acquired in
routine clinical care.

\section{Evaluation Metrics}
\label{sec:metrics}

\subsection{Classification Metrics}

Consider a test set of $N$ scored 30-second epochs, pooled over all of its
subjects. Each epoch carries a ground-truth stage and a predicted stage drawn
from the five AASM classes $\mathcal{C} = \{\mathrm{W}, \mathrm{N1},
\mathrm{N2}, \mathrm{N3}, \mathrm{R}\}$. Let $n_{ij}$ denote the number of
epochs whose ground-truth stage is $i$ and whose predicted stage is $j$, and let
$n_{i+}$ and $n_{+j}$ denote the row and column totals of this
confusion matrix. All three metrics are computed once over the pooled epochs of
a dataset; they are not averaged over per-subject values.

Accuracy is the proportion of epochs assigned the correct stage,
\begin{equation}
\mathrm{Acc} = \frac{1}{N} \sum_{c \in \mathcal{C}} n_{cc}.
\end{equation}
Precision and recall for stage $c$ are $P_c = n_{cc} / n_{+c}$ and
$R_c = n_{cc} / n_{c+}$, and macro-F1 averages their harmonic mean over the
five stages without weighting,
\begin{equation}
\mathrm{MF1} = \frac{1}{5} \sum_{c \in \mathcal{C}}
  \frac{2 P_c R_c}{P_c + R_c}.
\end{equation}
Cohen's kappa compares the observed agreement $p_o$ with the agreement $p_e$
expected from the marginals alone,
\begin{equation}
\kappa = \frac{p_o - p_e}{1 - p_e}, \qquad
p_o = \frac{1}{N} \sum_{c \in \mathcal{C}} n_{cc}, \qquad
p_e = \frac{1}{N^{2}} \sum_{c \in \mathcal{C}} n_{c+}\, n_{+c},
\end{equation}
and is used in its unweighted form. The three are reported together because
they summarise different properties of the same confusion matrix: accuracy is
dominated by the majority stages, macro-F1 gives every stage equal weight, and
$\kappa$ discounts the agreement expected from the marginals.

The Avg column of Table~2 in the main paper is the unweighted mean of the four
per-dataset values.

The main paper reports point estimates. Ninety-five per cent confidence
intervals are obtained from 1,000 subject-level cluster bootstrap resamples, in
which subjects rather than epochs are drawn with replacement. Intervals for
every reported value are given in Section~\ref{sec:perclass}, together with
per-class F1 for the five sleep stages.

\subsection{Reasoning Quality Metrics}

\paragraph{Mention rate.}
The automated audit checks every rationale against the stage-defining
propositions triggered by its ground-truth stage. The LLM judge returns a yes,
no or ambiguous verdict for each triggered proposition, and ambiguous verdicts
are excluded from both the numerator and the denominator. The mention rate of a
proposition is the number of yes verdicts divided by the number of decisive
verdicts,
\begin{equation}
\text{mention rate} =
  \frac{n_{\mathrm{yes}}}{n_{\mathrm{yes}} + n_{\mathrm{no}}},
\end{equation}
taken over all audited epochs whose ground-truth stage triggers that
proposition. Confidence intervals use the same subject-level cluster bootstrap
as the classification metrics. The proposition inventory, the judge prompts and
the per-proposition results are given in Section~\ref{sec:audit}.

\paragraph{Expert rating.}
Two trained sleep technologists score each sampled rationale on three
dimensions: Factual Accuracy \& Perceptual Fidelity, Diagnostic Evidence
Comprehensiveness, and Logical Coherence \& Guideline Concordance, shortened in
the main paper to factual accuracy, evidence comprehensiveness and logical
coherence. Each dimension is scored independently on an integer scale from 0 to
5, with 5 the highest. Each sample therefore receives two ratings per dimension,
one from each rater, and the reported score distributions are over these
ratings. Agreement between the raters is quantified by
the intraclass correlation coefficient in its two-way random-effects,
absolute-agreement, average-measures form, ICC(2,k)
\citep{koo2016iccguideline}, pooled across all paired ratings. The full rating
scale is given in Section~\ref{sec:ratings}.

\section{Experimental Setup}
\label{sec:setup}

\noindent
\paragraph{Main experiments.}
The 62 MASS-EX (SS3) subjects were split at the subject level. Supervised
fine-tuning used subjects 01-03-0001 to 01-03-0052, of which the first five
carry fine-grained annotations and the other 45 carry coarse annotations;
subjects 01-03-0053 to 01-03-0064 formed the validation set, and the 53 subjects
of MASS-SS1 were the held-out test set. Each configuration was trained three
times, and the tables in the main paper report the mean over the three runs.

\paragraph{Efficiency measurement.}
The model size in Table~4 of the main paper is the disk size of all weight and
configuration files in the model directory. Throughput was measured on staging
requests at a concurrency of 256, using the inference settings given in the main
paper. Throughput counts epochs completed per second, and excludes waveform
rendering and output parsing.

\section{Detailed Results}
\label{sec:perclass}

\noindent
Each entry gives the point estimate with its 95\% confidence interval below,
obtained as described in Section~\ref{sec:metrics}.

{\scriptsize
\setlength{\tabcolsep}{2pt}
\setlength{\LTleft}{0pt}\setlength{\LTright}{0pt}
\begin{longtable}{@{\extracolsep{\fill}}l*{8}{c}@{}}
\caption{Baseline comparison on the four evaluation datasets.}
\label{tab:detmain}\\
\toprule
\textbf{Method} & \textbf{Acc} & \textbf{MF1} & $\boldsymbol{\kappa}$ & \textbf{F1-W} & \textbf{F1-N1} & \textbf{F1-N2} & \textbf{F1-N3} & \textbf{F1-R} \\
\midrule
\endfirsthead
\multicolumn{9}{@{}l}{\footnotesize\itshape Table \thetable\ (continued)}\\
\toprule
\textbf{Method} & \textbf{Acc} & \textbf{MF1} & $\boldsymbol{\kappa}$ & \textbf{F1-W} & \textbf{F1-N1} & \textbf{F1-N2} & \textbf{F1-N3} & \textbf{F1-R} \\
\midrule
\endhead
\midrule
\endfoot
\bottomrule
\endlastfoot
\multicolumn{9}{@{}l}{\textit{MASS-SS1} ($n = 53$)}\\
\midrule
AttnSleep & 0.801 & 0.756 & 0.723 & 0.912 & 0.514 & 0.841 & 0.712 & 0.801\\*[-2pt]
 & {\tiny [0.779, 0.821]} & {\tiny [0.731, 0.777]} & {\tiny [0.693, 0.750]} & {\tiny [0.890, 0.929]} & {\tiny [0.483, 0.545]} & {\tiny [0.824, 0.857]} & {\tiny [0.657, 0.757]} & {\tiny [0.749, 0.840]}\\\addlinespace[2.5pt]
DeepSleepNet & 0.803 & 0.751 & 0.725 & 0.907 & 0.558 & 0.840 & 0.606 & 0.844\\*[-2pt]
 & {\tiny [0.783, 0.821]} & {\tiny [0.724, 0.773]} & {\tiny [0.696, 0.750]} & {\tiny [0.883, 0.926]} & {\tiny [0.528, 0.586]} & {\tiny [0.821, 0.857]} & {\tiny [0.522, 0.663]} & {\tiny [0.796, 0.883]}\\\addlinespace[2.5pt]
LPSGM & 0.831 & 0.794 & 0.763 & 0.913 & 0.600 & 0.867 & 0.715 & 0.876\\*[-2pt]
 & {\tiny [0.817, 0.843]} & {\tiny [0.777, 0.809]} & {\tiny [0.745, 0.781]} & {\tiny [0.896, 0.927]} & {\tiny [0.576, 0.625]} & {\tiny [0.854, 0.880]} & {\tiny [0.663, 0.756]} & {\tiny [0.846, 0.901]}\\\addlinespace[2.5pt]
ResnetMHA & 0.784 & 0.761 & 0.709 & 0.907 & 0.561 & 0.807 & 0.679 & 0.848\\*[-2pt]
 & {\tiny [0.758, 0.807]} & {\tiny [0.729, 0.787]} & {\tiny [0.675, 0.739]} & {\tiny [0.887, 0.923]} & {\tiny [0.531, 0.589]} & {\tiny [0.782, 0.828]} & {\tiny [0.595, 0.744]} & {\tiny [0.799, 0.885]}\\\addlinespace[2.5pt]
RobustSleepNet & 0.815 & 0.755 & 0.738 & 0.903 & 0.586 & 0.855 & 0.589 & 0.844\\*[-2pt]
 & {\tiny [0.800, 0.829]} & {\tiny [0.733, 0.774]} & {\tiny [0.718, 0.758]} & {\tiny [0.886, 0.917]} & {\tiny [0.557, 0.615]} & {\tiny [0.837, 0.871]} & {\tiny [0.499, 0.657]} & {\tiny [0.815, 0.870]}\\\addlinespace[2.5pt]
SalientSleepNet & 0.787 & 0.722 & 0.699 & 0.876 & 0.477 & 0.849 & 0.612 & 0.797\\*[-2pt]
 & {\tiny [0.770, 0.804]} & {\tiny [0.701, 0.741]} & {\tiny [0.675, 0.721]} & {\tiny [0.853, 0.897]} & {\tiny [0.453, 0.499]} & {\tiny [0.834, 0.862]} & {\tiny [0.544, 0.666]} & {\tiny [0.754, 0.834]}\\\addlinespace[2.5pt]
SeqSleepNet & 0.799 & 0.742 & 0.712 & 0.886 & 0.465 & 0.851 & 0.673 & 0.835\\*[-2pt]
 & {\tiny [0.782, 0.815]} & {\tiny [0.720, 0.761]} & {\tiny [0.684, 0.736]} & {\tiny [0.864, 0.905]} & {\tiny [0.438, 0.494]} & {\tiny [0.837, 0.865]} & {\tiny [0.609, 0.724]} & {\tiny [0.792, 0.871]}\\\addlinespace[2.5pt]
SleepDG & 0.812 & 0.762 & 0.735 & 0.888 & 0.514 & 0.860 & 0.701 & 0.849\\*[-2pt]
 & {\tiny [0.794, 0.827]} & {\tiny [0.742, 0.779]} & {\tiny [0.709, 0.756]} & {\tiny [0.862, 0.910]} & {\tiny [0.482, 0.543]} & {\tiny [0.845, 0.874]} & {\tiny [0.642, 0.752]} & {\tiny [0.818, 0.873]}\\\addlinespace[2.5pt]
TinySleepNet & 0.829 & 0.779 & 0.760 & 0.909 & 0.568 & 0.873 & 0.692 & 0.855\\*[-2pt]
 & {\tiny [0.811, 0.845]} & {\tiny [0.758, 0.799]} & {\tiny [0.735, 0.782]} & {\tiny [0.891, 0.925]} & {\tiny [0.539, 0.597]} & {\tiny [0.859, 0.886]} & {\tiny [0.632, 0.743]} & {\tiny [0.822, 0.880]}\\\addlinespace[2.5pt]
U-Sleep & 0.812 & 0.749 & 0.732 & 0.885 & 0.505 & 0.862 & 0.632 & 0.858\\*[-2pt]
 & {\tiny [0.795, 0.827]} & {\tiny [0.727, 0.766]} & {\tiny [0.708, 0.753]} & {\tiny [0.862, 0.905]} & {\tiny [0.479, 0.531]} & {\tiny [0.848, 0.875]} & {\tiny [0.570, 0.682]} & {\tiny [0.827, 0.882]}\\\addlinespace[2.5pt]
U-Time & 0.770 & 0.703 & 0.675 & 0.859 & 0.440 & 0.828 & 0.599 & 0.787\\*[-2pt]
 & {\tiny [0.748, 0.790]} & {\tiny [0.676, 0.726]} & {\tiny [0.643, 0.703]} & {\tiny [0.824, 0.887]} & {\tiny [0.413, 0.467]} & {\tiny [0.812, 0.842]} & {\tiny [0.523, 0.661]} & {\tiny [0.732, 0.830]}\\\addlinespace[2.5pt]
XSleepNet & 0.780 & 0.711 & 0.689 & 0.853 & 0.430 & 0.841 & 0.607 & 0.825\\*[-2pt]
 & {\tiny [0.756, 0.802]} & {\tiny [0.684, 0.737]} & {\tiny [0.656, 0.718]} & {\tiny [0.824, 0.879]} & {\tiny [0.397, 0.467]} & {\tiny [0.820, 0.860]} & {\tiny [0.524, 0.677]} & {\tiny [0.789, 0.859]}\\\addlinespace[2.5pt]
SleepXViT & 0.838 & 0.793 & 0.771 & 0.917 & 0.543 & 0.877 & 0.752 & 0.878\\*[-2pt]
 & {\tiny [0.825, 0.850]} & {\tiny [0.775, 0.809]} & {\tiny [0.752, 0.788]} & {\tiny [0.904, 0.929]} & {\tiny [0.511, 0.572]} & {\tiny [0.867, 0.888]} & {\tiny [0.696, 0.800]} & {\tiny [0.847, 0.900]}\\\addlinespace[2.5pt]
Resnet-18 & 0.830 & 0.778 & 0.758 & 0.918 & 0.523 & 0.873 & 0.737 & 0.839\\*[-2pt]
 & {\tiny [0.815, 0.843]} & {\tiny [0.760, 0.793]} & {\tiny [0.737, 0.778]} & {\tiny [0.901, 0.932]} & {\tiny [0.486, 0.559]} & {\tiny [0.861, 0.884]} & {\tiny [0.686, 0.778]} & {\tiny [0.811, 0.863]}\\\addlinespace[2.5pt]
\textbf{SleepVLM (ours)} & 0.835 & 0.793 & 0.767 & 0.917 & 0.527 & 0.872 & 0.784 & 0.863\\*[-2pt]
 & {\tiny [0.824, 0.846]} & {\tiny [0.777, 0.806]} & {\tiny [0.750, 0.782]} & {\tiny [0.904, 0.929]} & {\tiny [0.498, 0.555]} & {\tiny [0.861, 0.884]} & {\tiny [0.722, 0.827]} & {\tiny [0.842, 0.880]}\\\addlinespace[2.5pt]
\midrule
\multicolumn{9}{@{}l}{\textit{ZUAMHCS} ($n = 100$)}\\
\midrule
AttnSleep & 0.752 & 0.701 & 0.666 & 0.754 & 0.401 & 0.817 & 0.809 & 0.724\\*[-2pt]
 & {\tiny [0.731, 0.771]} & {\tiny [0.679, 0.722]} & {\tiny [0.640, 0.690]} & {\tiny [0.701, 0.799]} & {\tiny [0.373, 0.429]} & {\tiny [0.798, 0.834]} & {\tiny [0.785, 0.830]} & {\tiny [0.694, 0.752]}\\\addlinespace[2.5pt]
DeepSleepNet & 0.749 & 0.701 & 0.658 & 0.769 & 0.443 & 0.816 & 0.758 & 0.720\\*[-2pt]
 & {\tiny [0.728, 0.768]} & {\tiny [0.676, 0.722]} & {\tiny [0.630, 0.683]} & {\tiny [0.708, 0.818]} & {\tiny [0.415, 0.470]} & {\tiny [0.797, 0.833]} & {\tiny [0.727, 0.783]} & {\tiny [0.684, 0.752]}\\\addlinespace[2.5pt]
LPSGM & 0.818 & 0.775 & 0.750 & 0.819 & 0.533 & 0.859 & 0.811 & 0.853\\*[-2pt]
 & {\tiny [0.797, 0.833]} & {\tiny [0.752, 0.793]} & {\tiny [0.722, 0.771]} & {\tiny [0.761, 0.863]} & {\tiny [0.506, 0.560]} & {\tiny [0.842, 0.874]} & {\tiny [0.785, 0.832]} & {\tiny [0.832, 0.871]}\\\addlinespace[2.5pt]
ResnetMHA & 0.751 & 0.712 & 0.667 & 0.827 & 0.451 & 0.790 & 0.803 & 0.688\\*[-2pt]
 & {\tiny [0.729, 0.772]} & {\tiny [0.686, 0.735]} & {\tiny [0.637, 0.694]} & {\tiny [0.763, 0.869]} & {\tiny [0.423, 0.477]} & {\tiny [0.769, 0.809]} & {\tiny [0.776, 0.826]} & {\tiny [0.645, 0.732]}\\\addlinespace[2.5pt]
RobustSleepNet & 0.799 & 0.757 & 0.725 & 0.823 & 0.527 & 0.836 & 0.791 & 0.809\\*[-2pt]
 & {\tiny [0.783, 0.813]} & {\tiny [0.741, 0.771]} & {\tiny [0.704, 0.743]} & {\tiny [0.790, 0.850]} & {\tiny [0.500, 0.553]} & {\tiny [0.818, 0.852]} & {\tiny [0.758, 0.818]} & {\tiny [0.780, 0.836]}\\\addlinespace[2.5pt]
SalientSleepNet & 0.778 & 0.732 & 0.693 & 0.789 & 0.468 & 0.827 & 0.738 & 0.837\\*[-2pt]
 & {\tiny [0.758, 0.795]} & {\tiny [0.710, 0.750]} & {\tiny [0.668, 0.716]} & {\tiny [0.735, 0.833]} & {\tiny [0.439, 0.495]} & {\tiny [0.809, 0.844]} & {\tiny [0.701, 0.770]} & {\tiny [0.811, 0.860]}\\\addlinespace[2.5pt]
SeqSleepNet & 0.812 & 0.757 & 0.737 & 0.839 & 0.435 & 0.850 & 0.811 & 0.850\\*[-2pt]
 & {\tiny [0.793, 0.827]} & {\tiny [0.739, 0.773]} & {\tiny [0.713, 0.758]} & {\tiny [0.813, 0.860]} & {\tiny [0.400, 0.468]} & {\tiny [0.832, 0.867]} & {\tiny [0.787, 0.831]} & {\tiny [0.826, 0.871]}\\\addlinespace[2.5pt]
SleepDG & 0.798 & 0.749 & 0.719 & 0.817 & 0.466 & 0.833 & 0.812 & 0.816\\*[-2pt]
 & {\tiny [0.777, 0.816]} & {\tiny [0.725, 0.768]} & {\tiny [0.690, 0.743]} & {\tiny [0.754, 0.864]} & {\tiny [0.437, 0.492]} & {\tiny [0.816, 0.849]} & {\tiny [0.784, 0.835]} & {\tiny [0.791, 0.839]}\\\addlinespace[2.5pt]
TinySleepNet & 0.786 & 0.741 & 0.712 & 0.799 & 0.448 & 0.819 & 0.838 & 0.800\\*[-2pt]
 & {\tiny [0.764, 0.806]} & {\tiny [0.718, 0.763]} & {\tiny [0.684, 0.738]} & {\tiny [0.758, 0.834]} & {\tiny [0.393, 0.495]} & {\tiny [0.801, 0.837]} & {\tiny [0.813, 0.860]} & {\tiny [0.775, 0.823]}\\\addlinespace[2.5pt]
U-Sleep & 0.781 & 0.720 & 0.697 & 0.783 & 0.407 & 0.834 & 0.777 & 0.797\\*[-2pt]
 & {\tiny [0.764, 0.797]} & {\tiny [0.702, 0.737]} & {\tiny [0.674, 0.718]} & {\tiny [0.737, 0.821]} & {\tiny [0.380, 0.436]} & {\tiny [0.818, 0.850]} & {\tiny [0.753, 0.799]} & {\tiny [0.772, 0.821]}\\\addlinespace[2.5pt]
U-Time & 0.776 & 0.715 & 0.691 & 0.728 & 0.425 & 0.839 & 0.818 & 0.763\\*[-2pt]
 & {\tiny [0.756, 0.793]} & {\tiny [0.693, 0.734]} & {\tiny [0.665, 0.714]} & {\tiny [0.669, 0.779]} & {\tiny [0.401, 0.449]} & {\tiny [0.822, 0.853]} & {\tiny [0.793, 0.839]} & {\tiny [0.728, 0.795]}\\\addlinespace[2.5pt]
XSleepNet & 0.727 & 0.661 & 0.634 & 0.694 & 0.316 & 0.806 & 0.789 & 0.701\\*[-2pt]
 & {\tiny [0.702, 0.751]} & {\tiny [0.637, 0.686]} & {\tiny [0.601, 0.665]} & {\tiny [0.644, 0.740]} & {\tiny [0.281, 0.351]} & {\tiny [0.783, 0.826]} & {\tiny [0.763, 0.812]} & {\tiny [0.649, 0.746]}\\\addlinespace[2.5pt]
SleepXViT & 0.776 & 0.724 & 0.694 & 0.814 & 0.428 & 0.830 & 0.839 & 0.711\\*[-2pt]
 & {\tiny [0.757, 0.795]} & {\tiny [0.705, 0.745]} & {\tiny [0.668, 0.719]} & {\tiny [0.774, 0.850]} & {\tiny [0.393, 0.459]} & {\tiny [0.812, 0.845]} & {\tiny [0.816, 0.858]} & {\tiny [0.674, 0.747]}\\\addlinespace[2.5pt]
Resnet-18 & 0.793 & 0.747 & 0.717 & 0.799 & 0.487 & 0.826 & 0.829 & 0.795\\*[-2pt]
 & {\tiny [0.771, 0.812]} & {\tiny [0.723, 0.768]} & {\tiny [0.687, 0.742]} & {\tiny [0.738, 0.850]} & {\tiny [0.457, 0.515]} & {\tiny [0.808, 0.842]} & {\tiny [0.802, 0.852]} & {\tiny [0.770, 0.818]}\\\addlinespace[2.5pt]
\textbf{SleepVLM (ours)} & 0.812 & 0.766 & 0.743 & 0.833 & 0.486 & 0.849 & 0.855 & 0.806\\*[-2pt]
 & {\tiny [0.796, 0.827]} & {\tiny [0.749, 0.782]} & {\tiny [0.721, 0.763]} & {\tiny [0.797, 0.861]} & {\tiny [0.456, 0.514]} & {\tiny [0.835, 0.863]} & {\tiny [0.834, 0.873]} & {\tiny [0.783, 0.828]}\\\addlinespace[2.5pt]
\midrule
\multicolumn{9}{@{}l}{\textit{ABC} ($n = 132$)}\\
\midrule
AttnSleep & 0.705 & 0.674 & 0.602 & 0.799 & 0.452 & 0.764 & 0.697 & 0.660\\*[-2pt]
 & {\tiny [0.684, 0.725]} & {\tiny [0.652, 0.693]} & {\tiny [0.574, 0.625]} & {\tiny [0.771, 0.827]} & {\tiny [0.428, 0.472]} & {\tiny [0.739, 0.785]} & {\tiny [0.658, 0.733]} & {\tiny [0.622, 0.693]}\\\addlinespace[2.5pt]
DeepSleepNet & 0.652 & 0.606 & 0.534 & 0.767 & 0.438 & 0.713 & 0.506 & 0.605\\*[-2pt]
 & {\tiny [0.629, 0.675]} & {\tiny [0.582, 0.629]} & {\tiny [0.504, 0.561]} & {\tiny [0.738, 0.795]} & {\tiny [0.414, 0.461]} & {\tiny [0.682, 0.739]} & {\tiny [0.453, 0.557]} & {\tiny [0.556, 0.649]}\\\addlinespace[2.5pt]
LPSGM & 0.683 & 0.663 & 0.580 & 0.780 & 0.495 & 0.718 & 0.567 & 0.754\\*[-2pt]
 & {\tiny [0.655, 0.707]} & {\tiny [0.636, 0.687]} & {\tiny [0.545, 0.609]} & {\tiny [0.752, 0.806]} & {\tiny [0.469, 0.519]} & {\tiny [0.680, 0.749]} & {\tiny [0.515, 0.618]} & {\tiny [0.712, 0.792]}\\\addlinespace[2.5pt]
ResnetMHA & 0.716 & 0.715 & 0.627 & 0.819 & 0.482 & 0.738 & 0.751 & 0.784\\*[-2pt]
 & {\tiny [0.696, 0.737]} & {\tiny [0.695, 0.734]} & {\tiny [0.602, 0.654]} & {\tiny [0.801, 0.839]} & {\tiny [0.460, 0.502]} & {\tiny [0.707, 0.765]} & {\tiny [0.709, 0.789]} & {\tiny [0.743, 0.817]}\\\addlinespace[2.5pt]
RobustSleepNet & 0.597 & 0.524 & 0.473 & 0.731 & 0.449 & 0.623 & 0.223 & 0.592\\*[-2pt]
 & {\tiny [0.571, 0.623]} & {\tiny [0.498, 0.546]} & {\tiny [0.441, 0.503]} & {\tiny [0.698, 0.762]} & {\tiny [0.422, 0.474]} & {\tiny [0.581, 0.661]} & {\tiny [0.166, 0.278]} & {\tiny [0.553, 0.627]}\\\addlinespace[2.5pt]
SalientSleepNet & 0.639 & 0.564 & 0.518 & 0.777 & 0.430 & 0.676 & 0.243 & 0.695\\*[-2pt]
 & {\tiny [0.616, 0.660]} & {\tiny [0.540, 0.586]} & {\tiny [0.491, 0.544]} & {\tiny [0.751, 0.802]} & {\tiny [0.407, 0.452]} & {\tiny [0.643, 0.707]} & {\tiny [0.175, 0.306]} & {\tiny [0.654, 0.735]}\\\addlinespace[2.5pt]
SeqSleepNet & 0.669 & 0.623 & 0.546 & 0.775 & 0.400 & 0.753 & 0.544 & 0.643\\*[-2pt]
 & {\tiny [0.647, 0.689]} & {\tiny [0.600, 0.644]} & {\tiny [0.518, 0.573]} & {\tiny [0.752, 0.797]} & {\tiny [0.374, 0.426]} & {\tiny [0.725, 0.775]} & {\tiny [0.487, 0.594]} & {\tiny [0.597, 0.683]}\\\addlinespace[2.5pt]
SleepDG & 0.729 & 0.698 & 0.631 & 0.757 & 0.410 & 0.796 & 0.737 & 0.790\\*[-2pt]
 & {\tiny [0.708, 0.748]} & {\tiny [0.680, 0.714]} & {\tiny [0.604, 0.655]} & {\tiny [0.727, 0.788]} & {\tiny [0.386, 0.434]} & {\tiny [0.775, 0.814]} & {\tiny [0.697, 0.771]} & {\tiny [0.760, 0.817]}\\\addlinespace[2.5pt]
TinySleepNet & 0.719 & 0.686 & 0.619 & 0.751 & 0.451 & 0.796 & 0.710 & 0.723\\*[-2pt]
 & {\tiny [0.695, 0.739]} & {\tiny [0.662, 0.706]} & {\tiny [0.588, 0.645]} & {\tiny [0.717, 0.781]} & {\tiny [0.424, 0.475]} & {\tiny [0.774, 0.815]} & {\tiny [0.667, 0.747]} & {\tiny [0.681, 0.758]}\\\addlinespace[2.5pt]
U-Sleep & 0.720 & 0.675 & 0.614 & 0.760 & 0.418 & 0.798 & 0.648 & 0.749\\*[-2pt]
 & {\tiny [0.701, 0.738]} & {\tiny [0.657, 0.690]} & {\tiny [0.588, 0.637]} & {\tiny [0.737, 0.785]} & {\tiny [0.397, 0.438]} & {\tiny [0.778, 0.817]} & {\tiny [0.604, 0.688]} & {\tiny [0.720, 0.777]}\\\addlinespace[2.5pt]
U-Time & 0.682 & 0.642 & 0.562 & 0.718 & 0.370 & 0.756 & 0.628 & 0.738\\*[-2pt]
 & {\tiny [0.664, 0.703]} & {\tiny [0.624, 0.659]} & {\tiny [0.537, 0.586]} & {\tiny [0.691, 0.742]} & {\tiny [0.347, 0.389]} & {\tiny [0.736, 0.775]} & {\tiny [0.583, 0.668]} & {\tiny [0.704, 0.770]}\\\addlinespace[2.5pt]
XSleepNet & 0.489 & 0.436 & 0.320 & 0.561 & 0.198 & 0.566 & 0.440 & 0.415\\*[-2pt]
 & {\tiny [0.462, 0.516]} & {\tiny [0.408, 0.463]} & {\tiny [0.286, 0.352]} & {\tiny [0.529, 0.593]} & {\tiny [0.171, 0.229]} & {\tiny [0.519, 0.605]} & {\tiny [0.377, 0.499]} & {\tiny [0.342, 0.475]}\\\addlinespace[2.5pt]
SleepXViT & 0.703 & 0.671 & 0.592 & 0.769 & 0.442 & 0.786 & 0.744 & 0.612\\*[-2pt]
 & {\tiny [0.685, 0.720]} & {\tiny [0.651, 0.687]} & {\tiny [0.567, 0.614]} & {\tiny [0.744, 0.794]} & {\tiny [0.420, 0.465]} & {\tiny [0.765, 0.804]} & {\tiny [0.708, 0.773]} & {\tiny [0.565, 0.651]}\\\addlinespace[2.5pt]
Resnet-18 & 0.713 & 0.689 & 0.609 & 0.768 & 0.462 & 0.781 & 0.735 & 0.699\\*[-2pt]
 & {\tiny [0.695, 0.731]} & {\tiny [0.670, 0.705]} & {\tiny [0.584, 0.632]} & {\tiny [0.740, 0.796]} & {\tiny [0.439, 0.484]} & {\tiny [0.760, 0.798]} & {\tiny [0.698, 0.766]} & {\tiny [0.662, 0.730]}\\\addlinespace[2.5pt]
\textbf{SleepVLM (ours)} & 0.739 & 0.706 & 0.640 & 0.804 & 0.409 & 0.803 & 0.735 & 0.777\\*[-2pt]
 & {\tiny [0.720, 0.757]} & {\tiny [0.688, 0.721]} & {\tiny [0.615, 0.663]} & {\tiny [0.782, 0.827]} & {\tiny [0.385, 0.434]} & {\tiny [0.782, 0.821]} & {\tiny [0.698, 0.768]} & {\tiny [0.745, 0.801]}\\\addlinespace[2.5pt]
\midrule
\multicolumn{9}{@{}l}{\textit{DCSM} ($n = 255$)}\\
\midrule
AttnSleep & 0.662 & 0.600 & 0.548 & 0.767 & 0.284 & 0.707 & 0.573 & 0.671\\*[-2pt]
 & {\tiny [0.644, 0.680]} & {\tiny [0.581, 0.618]} & {\tiny [0.524, 0.571]} & {\tiny [0.744, 0.788]} & {\tiny [0.262, 0.305]} & {\tiny [0.686, 0.726]} & {\tiny [0.533, 0.609]} & {\tiny [0.641, 0.697]}\\\addlinespace[2.5pt]
DeepSleepNet & 0.660 & 0.594 & 0.544 & 0.749 & 0.280 & 0.713 & 0.515 & 0.713\\*[-2pt]
 & {\tiny [0.640, 0.680]} & {\tiny [0.573, 0.614]} & {\tiny [0.518, 0.570]} & {\tiny [0.724, 0.774]} & {\tiny [0.257, 0.301]} & {\tiny [0.691, 0.734]} & {\tiny [0.471, 0.552]} & {\tiny [0.678, 0.743]}\\\addlinespace[2.5pt]
LPSGM & 0.657 & 0.616 & 0.552 & 0.767 & 0.315 & 0.694 & 0.584 & 0.720\\*[-2pt]
 & {\tiny [0.638, 0.676]} & {\tiny [0.597, 0.633]} & {\tiny [0.528, 0.575]} & {\tiny [0.749, 0.787]} & {\tiny [0.290, 0.339]} & {\tiny [0.671, 0.716]} & {\tiny [0.545, 0.620]} & {\tiny [0.688, 0.747]}\\\addlinespace[2.5pt]
ResnetMHA & 0.667 & 0.633 & 0.567 & 0.787 & 0.339 & 0.692 & 0.622 & 0.725\\*[-2pt]
 & {\tiny [0.648, 0.683]} & {\tiny [0.615, 0.649]} & {\tiny [0.544, 0.588]} & {\tiny [0.766, 0.807]} & {\tiny [0.317, 0.362]} & {\tiny [0.670, 0.712]} & {\tiny [0.586, 0.653]} & {\tiny [0.694, 0.754]}\\\addlinespace[2.5pt]
RobustSleepNet & 0.682 & 0.627 & 0.578 & 0.757 & 0.327 & 0.723 & 0.640 & 0.687\\*[-2pt]
 & {\tiny [0.663, 0.700]} & {\tiny [0.609, 0.644]} & {\tiny [0.553, 0.601]} & {\tiny [0.734, 0.779]} & {\tiny [0.306, 0.350]} & {\tiny [0.702, 0.744]} & {\tiny [0.609, 0.669]} & {\tiny [0.656, 0.715]}\\\addlinespace[2.5pt]
SalientSleepNet & 0.632 & 0.561 & 0.513 & 0.739 & 0.302 & 0.694 & 0.433 & 0.637\\*[-2pt]
 & {\tiny [0.613, 0.652]} & {\tiny [0.542, 0.579]} & {\tiny [0.490, 0.537]} & {\tiny [0.717, 0.761]} & {\tiny [0.279, 0.321]} & {\tiny [0.672, 0.716]} & {\tiny [0.390, 0.473]} & {\tiny [0.602, 0.671]}\\\addlinespace[2.5pt]
SeqSleepNet & 0.663 & 0.601 & 0.549 & 0.778 & 0.315 & 0.718 & 0.525 & 0.670\\*[-2pt]
 & {\tiny [0.643, 0.683]} & {\tiny [0.580, 0.620]} & {\tiny [0.523, 0.573]} & {\tiny [0.761, 0.795]} & {\tiny [0.293, 0.336]} & {\tiny [0.698, 0.738]} & {\tiny [0.481, 0.565]} & {\tiny [0.630, 0.708]}\\\addlinespace[2.5pt]
SleepDG & 0.692 & 0.631 & 0.587 & 0.786 & 0.349 & 0.742 & 0.546 & 0.732\\*[-2pt]
 & {\tiny [0.670, 0.710]} & {\tiny [0.611, 0.649]} & {\tiny [0.560, 0.610]} & {\tiny [0.764, 0.807]} & {\tiny [0.323, 0.375]} & {\tiny [0.723, 0.760]} & {\tiny [0.505, 0.584]} & {\tiny [0.696, 0.761]}\\\addlinespace[2.5pt]
TinySleepNet & 0.677 & 0.616 & 0.567 & 0.767 & 0.346 & 0.735 & 0.540 & 0.693\\*[-2pt]
 & {\tiny [0.657, 0.695]} & {\tiny [0.596, 0.634]} & {\tiny [0.542, 0.591]} & {\tiny [0.749, 0.785]} & {\tiny [0.317, 0.372]} & {\tiny [0.715, 0.753]} & {\tiny [0.496, 0.577]} & {\tiny [0.663, 0.719]}\\\addlinespace[2.5pt]
U-Sleep & 0.686 & 0.620 & 0.576 & 0.759 & 0.339 & 0.743 & 0.525 & 0.732\\*[-2pt]
 & {\tiny [0.669, 0.701]} & {\tiny [0.603, 0.635]} & {\tiny [0.555, 0.596]} & {\tiny [0.742, 0.777]} & {\tiny [0.319, 0.359]} & {\tiny [0.725, 0.758]} & {\tiny [0.487, 0.559]} & {\tiny [0.707, 0.754]}\\\addlinespace[2.5pt]
U-Time & 0.663 & 0.587 & 0.539 & 0.708 & 0.346 & 0.746 & 0.465 & 0.670\\*[-2pt]
 & {\tiny [0.647, 0.679]} & {\tiny [0.568, 0.603]} & {\tiny [0.518, 0.559]} & {\tiny [0.685, 0.730]} & {\tiny [0.326, 0.366]} & {\tiny [0.731, 0.760]} & {\tiny [0.420, 0.504]} & {\tiny [0.642, 0.695]}\\\addlinespace[2.5pt]
XSleepNet & 0.652 & 0.563 & 0.532 & 0.794 & 0.180 & 0.690 & 0.523 & 0.629\\*[-2pt]
 & {\tiny [0.633, 0.670]} & {\tiny [0.543, 0.582]} & {\tiny [0.506, 0.556]} & {\tiny [0.775, 0.813]} & {\tiny [0.161, 0.199]} & {\tiny [0.667, 0.712]} & {\tiny [0.477, 0.564]} & {\tiny [0.593, 0.662]}\\\addlinespace[2.5pt]
SleepXViT & 0.786 & 0.728 & 0.712 & 0.873 & 0.380 & 0.808 & 0.779 & 0.799\\*[-2pt]
 & {\tiny [0.774, 0.797]} & {\tiny [0.715, 0.739]} & {\tiny [0.696, 0.727]} & {\tiny [0.859, 0.885]} & {\tiny [0.360, 0.401]} & {\tiny [0.794, 0.820]} & {\tiny [0.759, 0.797]} & {\tiny [0.776, 0.820]}\\\addlinespace[2.5pt]
Resnet-18 & 0.781 & 0.716 & 0.705 & 0.836 & 0.323 & 0.817 & 0.776 & 0.829\\*[-2pt]
 & {\tiny [0.769, 0.792]} & {\tiny [0.705, 0.727]} & {\tiny [0.690, 0.720]} & {\tiny [0.823, 0.849]} & {\tiny [0.305, 0.343]} & {\tiny [0.805, 0.829]} & {\tiny [0.754, 0.794]} & {\tiny [0.810, 0.846]}\\\addlinespace[2.5pt]
\textbf{SleepVLM (ours)} & 0.789 & 0.733 & 0.717 & 0.882 & 0.403 & 0.800 & 0.775 & 0.805\\*[-2pt]
 & {\tiny [0.778, 0.800]} & {\tiny [0.720, 0.743]} & {\tiny [0.702, 0.730]} & {\tiny [0.870, 0.893]} & {\tiny [0.378, 0.427]} & {\tiny [0.787, 0.811]} & {\tiny [0.753, 0.794]} & {\tiny [0.785, 0.822]}\\\addlinespace[2.5pt]
\end{longtable}
\par}

{\scriptsize
\setlength{\tabcolsep}{2pt}
\setlength{\LTleft}{0pt}\setlength{\LTright}{0pt}
\begin{longtable}{@{\extracolsep{\fill}}l*{8}{c}@{}}
\caption{Ablation study on the four evaluation datasets.}
\label{tab:detabl}\\
\toprule
\textbf{Configuration} & \textbf{Acc} & \textbf{MF1} & $\boldsymbol{\kappa}$ & \textbf{F1-W} & \textbf{F1-N1} & \textbf{F1-N2} & \textbf{F1-N3} & \textbf{F1-R} \\
\midrule
\endfirsthead
\multicolumn{9}{@{}l}{\footnotesize\itshape Table \thetable\ (continued)}\\
\toprule
\textbf{Configuration} & \textbf{Acc} & \textbf{MF1} & $\boldsymbol{\kappa}$ & \textbf{F1-W} & \textbf{F1-N1} & \textbf{F1-N2} & \textbf{F1-N3} & \textbf{F1-R} \\
\midrule
\endhead
\midrule
\endfoot
\bottomrule
\endlastfoot
\multicolumn{9}{@{}l}{\textit{MASS-SS1} ($n = 53$)}\\
\midrule
Zero-shot baseline & 0.100 & 0.070 & $-$0.003 & 0.005 & 0.077 & 0.140 & 0.122 & 0.005\\*[-2pt]
 & {\tiny [0.087, 0.114]} & {\tiny [0.064, 0.076]} & {\tiny [$-$0.006, $-$0.001]} & {\tiny [0.003, 0.007]} & {\tiny [0.069, 0.083]} & {\tiny [0.131, 0.149]} & {\tiny [0.093, 0.152]} & {\tiny [0.003, 0.007]}\\\addlinespace[2.5pt]
\textbf{SleepVLM} & 0.835 & 0.793 & 0.767 & 0.917 & 0.527 & 0.872 & 0.784 & 0.863\\*[-2pt]
 & {\tiny [0.824, 0.846]} & {\tiny [0.777, 0.806]} & {\tiny [0.750, 0.782]} & {\tiny [0.904, 0.929]} & {\tiny [0.498, 0.555]} & {\tiny [0.861, 0.884]} & {\tiny [0.722, 0.827]} & {\tiny [0.842, 0.880]}\\\addlinespace[2.5pt]
w/o WPT & 0.818 & 0.753 & 0.736 & 0.910 & 0.380 & 0.858 & 0.766 & 0.853\\*[-2pt]
 & {\tiny [0.804, 0.831]} & {\tiny [0.735, 0.770]} & {\tiny [0.715, 0.755]} & {\tiny [0.896, 0.922]} & {\tiny [0.338, 0.423]} & {\tiny [0.846, 0.869]} & {\tiny [0.698, 0.817]} & {\tiny [0.829, 0.874]}\\\addlinespace[2.5pt]
w/o Rule Grounding & 0.829 & 0.785 & 0.758 & 0.902 & 0.530 & 0.877 & 0.765 & 0.852\\*[-2pt]
 & {\tiny [0.817, 0.840]} & {\tiny [0.768, 0.798]} & {\tiny [0.740, 0.773]} & {\tiny [0.885, 0.916]} & {\tiny [0.501, 0.557]} & {\tiny [0.866, 0.887]} & {\tiny [0.703, 0.810]} & {\tiny [0.826, 0.872]}\\\addlinespace[2.5pt]
+RFT & 0.810 & 0.738 & 0.728 & 0.892 & 0.354 & 0.863 & 0.772 & 0.809\\*[-2pt]
 & {\tiny [0.794, 0.824]} & {\tiny [0.718, 0.755]} & {\tiny [0.705, 0.749]} & {\tiny [0.874, 0.908]} & {\tiny [0.312, 0.394]} & {\tiny [0.851, 0.874]} & {\tiny [0.713, 0.817]} & {\tiny [0.775, 0.837]}\\\addlinespace[2.5pt]
w/o Coarse Annotation & 0.767 & 0.700 & 0.667 & 0.819 & 0.272 & 0.838 & 0.769 & 0.801\\*[-2pt]
 & {\tiny [0.749, 0.782]} & {\tiny [0.685, 0.711]} & {\tiny [0.641, 0.690]} & {\tiny [0.783, 0.847]} & {\tiny [0.234, 0.308]} & {\tiny [0.823, 0.851]} & {\tiny [0.716, 0.806]} & {\tiny [0.773, 0.826]}\\\addlinespace[2.5pt]
w/o Coarse Annot.\ \& WPT & 0.673 & 0.584 & 0.526 & 0.578 & 0.198 & 0.812 & 0.709 & 0.625\\*[-2pt]
 & {\tiny [0.649, 0.694]} & {\tiny [0.562, 0.603]} & {\tiny [0.497, 0.553]} & {\tiny [0.507, 0.637]} & {\tiny [0.177, 0.218]} & {\tiny [0.797, 0.826]} & {\tiny [0.643, 0.761]} & {\tiny [0.574, 0.668]}\\\addlinespace[2.5pt]
\midrule
\multicolumn{9}{@{}l}{\textit{ZUAMHCS} ($n = 100$)}\\
\midrule
Zero-shot baseline & 0.201 & 0.104 & $-$0.013 & 0.009 & 0.069 & 0.116 & 0.319 & 0.005\\*[-2pt]
 & {\tiny [0.186, 0.216]} & {\tiny [0.098, 0.108]} & {\tiny [$-$0.016, $-$0.010]} & {\tiny [0.007, 0.011]} & {\tiny [0.063, 0.076]} & {\tiny [0.112, 0.121]} & {\tiny [0.293, 0.344]} & {\tiny [0.003, 0.006]}\\\addlinespace[2.5pt]
\textbf{SleepVLM} & 0.812 & 0.766 & 0.743 & 0.833 & 0.486 & 0.849 & 0.855 & 0.806\\*[-2pt]
 & {\tiny [0.796, 0.827]} & {\tiny [0.749, 0.782]} & {\tiny [0.721, 0.763]} & {\tiny [0.797, 0.861]} & {\tiny [0.456, 0.514]} & {\tiny [0.835, 0.863]} & {\tiny [0.834, 0.873]} & {\tiny [0.783, 0.828]}\\\addlinespace[2.5pt]
w/o WPT & 0.804 & 0.743 & 0.727 & 0.824 & 0.411 & 0.837 & 0.848 & 0.796\\*[-2pt]
 & {\tiny [0.791, 0.818]} & {\tiny [0.728, 0.758]} & {\tiny [0.708, 0.747]} & {\tiny [0.789, 0.854]} & {\tiny [0.379, 0.442]} & {\tiny [0.823, 0.851]} & {\tiny [0.826, 0.867]} & {\tiny [0.775, 0.817]}\\\addlinespace[2.5pt]
w/o Rule Grounding & 0.795 & 0.746 & 0.720 & 0.828 & 0.438 & 0.838 & 0.846 & 0.777\\*[-2pt]
 & {\tiny [0.777, 0.813]} & {\tiny [0.725, 0.765]} & {\tiny [0.695, 0.744]} & {\tiny [0.790, 0.860]} & {\tiny [0.404, 0.473]} & {\tiny [0.821, 0.854]} & {\tiny [0.825, 0.865]} & {\tiny [0.751, 0.803]}\\\addlinespace[2.5pt]
+RFT & 0.801 & 0.737 & 0.727 & 0.802 & 0.392 & 0.830 & 0.840 & 0.818\\*[-2pt]
 & {\tiny [0.779, 0.820]} & {\tiny [0.712, 0.756]} & {\tiny [0.699, 0.753]} & {\tiny [0.740, 0.851]} & {\tiny [0.357, 0.424]} & {\tiny [0.812, 0.846]} & {\tiny [0.817, 0.859]} & {\tiny [0.793, 0.840]}\\\addlinespace[2.5pt]
w/o Coarse Annotation & 0.760 & 0.680 & 0.668 & 0.735 & 0.270 & 0.825 & 0.832 & 0.737\\*[-2pt]
 & {\tiny [0.740, 0.779]} & {\tiny [0.660, 0.699]} & {\tiny [0.641, 0.694]} & {\tiny [0.693, 0.773]} & {\tiny [0.234, 0.305]} & {\tiny [0.807, 0.842]} & {\tiny [0.809, 0.851]} & {\tiny [0.708, 0.765]}\\\addlinespace[2.5pt]
w/o Coarse Annot.\ \& WPT & 0.699 & 0.571 & 0.565 & 0.406 & 0.204 & 0.788 & 0.812 & 0.647\\*[-2pt]
 & {\tiny [0.680, 0.716]} & {\tiny [0.553, 0.587]} & {\tiny [0.540, 0.587]} & {\tiny [0.353, 0.453]} & {\tiny [0.177, 0.229]} & {\tiny [0.771, 0.805]} & {\tiny [0.786, 0.835]} & {\tiny [0.611, 0.682]}\\\addlinespace[2.5pt]
\midrule
\multicolumn{9}{@{}l}{\textit{ABC} ($n = 132$)}\\
\midrule
Zero-shot baseline & 0.093 & 0.120 & 0.008 & 0.027 & 0.105 & 0.174 & 0.162 & 0.133\\*[-2pt]
 & {\tiny [0.090, 0.096]} & {\tiny [0.117, 0.123]} & {\tiny [0.006, 0.010]} & {\tiny [0.024, 0.030]} & {\tiny [0.098, 0.112]} & {\tiny [0.168, 0.178]} & {\tiny [0.149, 0.176]} & {\tiny [0.126, 0.139]}\\\addlinespace[2.5pt]
\textbf{SleepVLM} & 0.739 & 0.706 & 0.640 & 0.804 & 0.409 & 0.803 & 0.735 & 0.777\\*[-2pt]
 & {\tiny [0.720, 0.757]} & {\tiny [0.688, 0.721]} & {\tiny [0.615, 0.663]} & {\tiny [0.782, 0.827]} & {\tiny [0.385, 0.434]} & {\tiny [0.782, 0.821]} & {\tiny [0.698, 0.768]} & {\tiny [0.745, 0.801]}\\\addlinespace[2.5pt]
w/o WPT & 0.687 & 0.656 & 0.575 & 0.730 & 0.341 & 0.762 & 0.720 & 0.726\\*[-2pt]
 & {\tiny [0.666, 0.706]} & {\tiny [0.637, 0.672]} & {\tiny [0.547, 0.599]} & {\tiny [0.706, 0.756]} & {\tiny [0.319, 0.361]} & {\tiny [0.739, 0.781]} & {\tiny [0.682, 0.753]} & {\tiny [0.695, 0.752]}\\\addlinespace[2.5pt]
w/o Rule Grounding & 0.704 & 0.651 & 0.581 & 0.755 & 0.331 & 0.795 & 0.701 & 0.673\\*[-2pt]
 & {\tiny [0.687, 0.719]} & {\tiny [0.634, 0.667]} & {\tiny [0.558, 0.603]} & {\tiny [0.734, 0.777]} & {\tiny [0.310, 0.351]} & {\tiny [0.777, 0.810]} & {\tiny [0.664, 0.735]} & {\tiny [0.632, 0.708]}\\\addlinespace[2.5pt]
+RFT & 0.681 & 0.605 & 0.547 & 0.714 & 0.261 & 0.790 & 0.668 & 0.591\\*[-2pt]
 & {\tiny [0.663, 0.696]} & {\tiny [0.587, 0.622]} & {\tiny [0.523, 0.570]} & {\tiny [0.688, 0.741]} & {\tiny [0.235, 0.287]} & {\tiny [0.773, 0.806]} & {\tiny [0.627, 0.704]} & {\tiny [0.544, 0.634]}\\\addlinespace[2.5pt]
w/o Coarse Annotation & 0.657 & 0.603 & 0.522 & 0.687 & 0.265 & 0.780 & 0.687 & 0.596\\*[-2pt]
 & {\tiny [0.639, 0.673]} & {\tiny [0.586, 0.619]} & {\tiny [0.497, 0.545]} & {\tiny [0.661, 0.714]} & {\tiny [0.245, 0.284]} & {\tiny [0.763, 0.796]} & {\tiny [0.648, 0.721]} & {\tiny [0.546, 0.639]}\\\addlinespace[2.5pt]
w/o Coarse Annot.\ \& WPT & 0.602 & 0.532 & 0.427 & 0.418 & 0.247 & 0.737 & 0.657 & 0.603\\*[-2pt]
 & {\tiny [0.580, 0.622]} & {\tiny [0.513, 0.549]} & {\tiny [0.399, 0.452]} & {\tiny [0.378, 0.466]} & {\tiny [0.226, 0.266]} & {\tiny [0.716, 0.755]} & {\tiny [0.609, 0.698]} & {\tiny [0.575, 0.629]}\\\addlinespace[2.5pt]
\midrule
\multicolumn{9}{@{}l}{\textit{DCSM} ($n = 255$)}\\
\midrule
Zero-shot baseline & 0.092 & 0.119 & 0.008 & 0.032 & 0.078 & 0.172 & 0.169 & 0.141\\*[-2pt]
 & {\tiny [0.090, 0.094]} & {\tiny [0.117, 0.121]} & {\tiny [0.007, 0.009]} & {\tiny [0.030, 0.034]} & {\tiny [0.073, 0.084]} & {\tiny [0.169, 0.176]} & {\tiny [0.162, 0.176]} & {\tiny [0.136, 0.146]}\\\addlinespace[2.5pt]
\textbf{SleepVLM} & 0.789 & 0.733 & 0.717 & 0.882 & 0.403 & 0.800 & 0.775 & 0.805\\*[-2pt]
 & {\tiny [0.778, 0.800]} & {\tiny [0.720, 0.743]} & {\tiny [0.702, 0.730]} & {\tiny [0.870, 0.893]} & {\tiny [0.378, 0.427]} & {\tiny [0.787, 0.811]} & {\tiny [0.753, 0.794]} & {\tiny [0.785, 0.822]}\\\addlinespace[2.5pt]
w/o WPT & 0.758 & 0.686 & 0.677 & 0.825 & 0.290 & 0.794 & 0.755 & 0.763\\*[-2pt]
 & {\tiny [0.746, 0.770]} & {\tiny [0.675, 0.697]} & {\tiny [0.661, 0.692]} & {\tiny [0.812, 0.838]} & {\tiny [0.273, 0.308]} & {\tiny [0.782, 0.807]} & {\tiny [0.734, 0.778]} & {\tiny [0.745, 0.780]}\\\addlinespace[2.5pt]
w/o Rule Grounding & 0.770 & 0.700 & 0.686 & 0.822 & 0.304 & 0.811 & 0.760 & 0.801\\*[-2pt]
 & {\tiny [0.759, 0.782]} & {\tiny [0.690, 0.711]} & {\tiny [0.671, 0.702]} & {\tiny [0.808, 0.836]} & {\tiny [0.287, 0.321]} & {\tiny [0.800, 0.822]} & {\tiny [0.741, 0.781]} & {\tiny [0.782, 0.821]}\\\addlinespace[2.5pt]
+RFT & 0.795 & 0.712 & 0.718 & 0.861 & 0.290 & 0.822 & 0.772 & 0.814\\*[-2pt]
 & {\tiny [0.785, 0.806]} & {\tiny [0.701, 0.722]} & {\tiny [0.704, 0.734]} & {\tiny [0.847, 0.874]} & {\tiny [0.269, 0.309]} & {\tiny [0.811, 0.833]} & {\tiny [0.751, 0.792]} & {\tiny [0.795, 0.834]}\\\addlinespace[2.5pt]
w/o Coarse Annotation & 0.732 & 0.661 & 0.636 & 0.774 & 0.255 & 0.796 & 0.713 & 0.767\\*[-2pt]
 & {\tiny [0.719, 0.744]} & {\tiny [0.648, 0.671]} & {\tiny [0.619, 0.651]} & {\tiny [0.757, 0.789]} & {\tiny [0.238, 0.271]} & {\tiny [0.784, 0.808]} & {\tiny [0.688, 0.736]} & {\tiny [0.747, 0.787]}\\\addlinespace[2.5pt]
w/o Coarse Annot.\ \& WPT & 0.616 & 0.538 & 0.476 & 0.424 & 0.230 & 0.746 & 0.731 & 0.562\\*[-2pt]
 & {\tiny [0.603, 0.628]} & {\tiny [0.527, 0.548]} & {\tiny [0.459, 0.491]} & {\tiny [0.404, 0.446]} & {\tiny [0.214, 0.244]} & {\tiny [0.731, 0.759]} & {\tiny [0.705, 0.753]} & {\tiny [0.542, 0.582]}\\\addlinespace[2.5pt]
\end{longtable}
\par}

{\scriptsize
\setlength{\tabcolsep}{2pt}
\setlength{\LTleft}{0pt}\setlength{\LTright}{0pt}
\begin{longtable}{@{\extracolsep{\fill}}l*{8}{c}@{}}
\caption{Quantization comparison on the four evaluation datasets.}
\label{tab:detquant}\\
\toprule
\textbf{Model} & \textbf{Acc} & \textbf{MF1} & $\boldsymbol{\kappa}$ & \textbf{F1-W} & \textbf{F1-N1} & \textbf{F1-N2} & \textbf{F1-N3} & \textbf{F1-R} \\
\midrule
\endfirsthead
\multicolumn{9}{@{}l}{\footnotesize\itshape Table \thetable\ (continued)}\\
\toprule
\textbf{Model} & \textbf{Acc} & \textbf{MF1} & $\boldsymbol{\kappa}$ & \textbf{F1-W} & \textbf{F1-N1} & \textbf{F1-N2} & \textbf{F1-N3} & \textbf{F1-R} \\
\midrule
\endhead
\midrule
\endfoot
\bottomrule
\endlastfoot
\multicolumn{9}{@{}l}{\textit{MASS-SS1} ($n = 53$)}\\
\midrule
Qwen2.5-VL-3B (BF16) & 0.835 & 0.793 & 0.767 & 0.917 & 0.527 & 0.872 & 0.784 & 0.863\\*[-2pt]
 & {\tiny [0.824, 0.846]} & {\tiny [0.777, 0.806]} & {\tiny [0.750, 0.782]} & {\tiny [0.904, 0.929]} & {\tiny [0.498, 0.555]} & {\tiny [0.861, 0.884]} & {\tiny [0.722, 0.827]} & {\tiny [0.842, 0.880]}\\\addlinespace[2.5pt]
Qwen2.5-VL-3B (W4A16) & 0.827 & 0.788 & 0.757 & 0.912 & 0.518 & 0.865 & 0.786 & 0.860\\*[-2pt]
 & {\tiny [0.815, 0.838]} & {\tiny [0.772, 0.801]} & {\tiny [0.740, 0.774]} & {\tiny [0.899, 0.924]} & {\tiny [0.490, 0.544]} & {\tiny [0.852, 0.877]} & {\tiny [0.724, 0.829]} & {\tiny [0.840, 0.876]}\\\addlinespace[2.5pt]
Qwen3.5-0.8B (BF16) & 0.832 & 0.773 & 0.761 & 0.918 & 0.454 & 0.874 & 0.745 & 0.875\\*[-2pt]
 & {\tiny [0.820, 0.844]} & {\tiny [0.756, 0.788]} & {\tiny [0.743, 0.777]} & {\tiny [0.905, 0.930]} & {\tiny [0.422, 0.485]} & {\tiny [0.863, 0.884]} & {\tiny [0.681, 0.793]} & {\tiny [0.855, 0.891]}\\\addlinespace[2.5pt]
Qwen3.5-0.8B (W4A16) & 0.832 & 0.774 & 0.760 & 0.915 & 0.462 & 0.875 & 0.743 & 0.873\\*[-2pt]
 & {\tiny [0.820, 0.843]} & {\tiny [0.757, 0.787]} & {\tiny [0.743, 0.777]} & {\tiny [0.901, 0.927]} & {\tiny [0.433, 0.488]} & {\tiny [0.865, 0.886]} & {\tiny [0.678, 0.794]} & {\tiny [0.851, 0.889]}\\\addlinespace[2.5pt]
\midrule
\multicolumn{9}{@{}l}{\textit{ZUAMHCS} ($n = 100$)}\\
\midrule
Qwen2.5-VL-3B (BF16) & 0.812 & 0.766 & 0.743 & 0.833 & 0.486 & 0.849 & 0.855 & 0.806\\*[-2pt]
 & {\tiny [0.796, 0.827]} & {\tiny [0.749, 0.782]} & {\tiny [0.721, 0.763]} & {\tiny [0.797, 0.861]} & {\tiny [0.456, 0.514]} & {\tiny [0.835, 0.863]} & {\tiny [0.834, 0.873]} & {\tiny [0.783, 0.828]}\\\addlinespace[2.5pt]
Qwen2.5-VL-3B (W4A16) & 0.798 & 0.751 & 0.727 & 0.827 & 0.443 & 0.837 & 0.854 & 0.796\\*[-2pt]
 & {\tiny [0.782, 0.813]} & {\tiny [0.733, 0.769]} & {\tiny [0.704, 0.748]} & {\tiny [0.789, 0.859]} & {\tiny [0.412, 0.472]} & {\tiny [0.822, 0.852]} & {\tiny [0.833, 0.871]} & {\tiny [0.772, 0.819]}\\\addlinespace[2.5pt]
Qwen3.5-0.8B (BF16) & 0.817 & 0.764 & 0.749 & 0.832 & 0.460 & 0.854 & 0.839 & 0.834\\*[-2pt]
 & {\tiny [0.798, 0.834]} & {\tiny [0.743, 0.782]} & {\tiny [0.723, 0.770]} & {\tiny [0.787, 0.867]} & {\tiny [0.426, 0.490]} & {\tiny [0.839, 0.867]} & {\tiny [0.815, 0.859]} & {\tiny [0.811, 0.855]}\\\addlinespace[2.5pt]
Qwen3.5-0.8B (W4A16) & 0.809 & 0.756 & 0.739 & 0.826 & 0.446 & 0.850 & 0.838 & 0.819\\*[-2pt]
 & {\tiny [0.789, 0.827]} & {\tiny [0.734, 0.775]} & {\tiny [0.712, 0.762]} & {\tiny [0.783, 0.861]} & {\tiny [0.412, 0.478]} & {\tiny [0.834, 0.865]} & {\tiny [0.814, 0.858]} & {\tiny [0.796, 0.841]}\\\addlinespace[2.5pt]
\midrule
\multicolumn{9}{@{}l}{\textit{ABC} ($n = 132$)}\\
\midrule
Qwen2.5-VL-3B (BF16) & 0.739 & 0.706 & 0.640 & 0.804 & 0.409 & 0.803 & 0.735 & 0.777\\*[-2pt]
 & {\tiny [0.720, 0.757]} & {\tiny [0.688, 0.721]} & {\tiny [0.615, 0.663]} & {\tiny [0.782, 0.827]} & {\tiny [0.385, 0.434]} & {\tiny [0.782, 0.821]} & {\tiny [0.698, 0.768]} & {\tiny [0.745, 0.801]}\\\addlinespace[2.5pt]
Qwen2.5-VL-3B (W4A16) & 0.728 & 0.697 & 0.627 & 0.796 & 0.403 & 0.796 & 0.732 & 0.756\\*[-2pt]
 & {\tiny [0.709, 0.746]} & {\tiny [0.679, 0.712]} & {\tiny [0.602, 0.651]} & {\tiny [0.773, 0.819]} & {\tiny [0.380, 0.426]} & {\tiny [0.773, 0.815]} & {\tiny [0.694, 0.764]} & {\tiny [0.722, 0.782]}\\\addlinespace[2.5pt]
Qwen3.5-0.8B (BF16) & 0.730 & 0.693 & 0.624 & 0.776 & 0.421 & 0.803 & 0.728 & 0.739\\*[-2pt]
 & {\tiny [0.711, 0.747]} & {\tiny [0.675, 0.709]} & {\tiny [0.600, 0.648]} & {\tiny [0.753, 0.799]} & {\tiny [0.397, 0.444]} & {\tiny [0.784, 0.820]} & {\tiny [0.688, 0.759]} & {\tiny [0.706, 0.765]}\\\addlinespace[2.5pt]
Qwen3.5-0.8B (W4A16) & 0.722 & 0.696 & 0.620 & 0.767 & 0.404 & 0.797 & 0.758 & 0.754\\*[-2pt]
 & {\tiny [0.703, 0.741]} & {\tiny [0.679, 0.711]} & {\tiny [0.594, 0.644]} & {\tiny [0.742, 0.791]} & {\tiny [0.380, 0.427]} & {\tiny [0.777, 0.815]} & {\tiny [0.725, 0.786]} & {\tiny [0.723, 0.779]}\\\addlinespace[2.5pt]
\midrule
\multicolumn{9}{@{}l}{\textit{DCSM} ($n = 255$)}\\
\midrule
Qwen2.5-VL-3B (BF16) & 0.789 & 0.733 & 0.717 & 0.882 & 0.403 & 0.800 & 0.775 & 0.805\\*[-2pt]
 & {\tiny [0.778, 0.800]} & {\tiny [0.720, 0.743]} & {\tiny [0.702, 0.730]} & {\tiny [0.870, 0.893]} & {\tiny [0.378, 0.427]} & {\tiny [0.787, 0.811]} & {\tiny [0.753, 0.794]} & {\tiny [0.785, 0.822]}\\\addlinespace[2.5pt]
Qwen2.5-VL-3B (W4A16) & 0.776 & 0.712 & 0.700 & 0.833 & 0.314 & 0.820 & 0.774 & 0.820\\*[-2pt]
 & {\tiny [0.765, 0.788]} & {\tiny [0.701, 0.722]} & {\tiny [0.685, 0.715]} & {\tiny [0.819, 0.847]} & {\tiny [0.296, 0.332]} & {\tiny [0.808, 0.831]} & {\tiny [0.752, 0.793]} & {\tiny [0.802, 0.837]}\\\addlinespace[2.5pt]
Qwen3.5-0.8B (BF16) & 0.775 & 0.716 & 0.698 & 0.827 & 0.363 & 0.817 & 0.763 & 0.811\\*[-2pt]
 & {\tiny [0.763, 0.787]} & {\tiny [0.704, 0.727]} & {\tiny [0.682, 0.713]} & {\tiny [0.811, 0.841]} & {\tiny [0.345, 0.382]} & {\tiny [0.805, 0.827]} & {\tiny [0.742, 0.782]} & {\tiny [0.791, 0.828]}\\\addlinespace[2.5pt]
Qwen3.5-0.8B (W4A16) & 0.765 & 0.706 & 0.687 & 0.823 & 0.326 & 0.809 & 0.770 & 0.803\\*[-2pt]
 & {\tiny [0.753, 0.777]} & {\tiny [0.694, 0.717]} & {\tiny [0.671, 0.703]} & {\tiny [0.808, 0.837]} & {\tiny [0.307, 0.345]} & {\tiny [0.797, 0.820]} & {\tiny [0.749, 0.788]} & {\tiny [0.783, 0.820]}\\\addlinespace[2.5pt]
\end{longtable}
\par}

\section{Automated Audit Details}
\label{sec:audit}

\noindent
Every rationale produced by SleepVLM is audited against the propositions
triggered by its ground-truth stage. The 11 required propositions (F1--F11 in
Table~\ref{tab:props}) are derived from the stage-onset rules of
Section~\ref{sec:rules}: rules W.1, W.2, N1.1 and N3.1 each contribute one
proposition matching the defining feature of the rule, and the multi-feature
rules N1.2, N2.1 and R.1 decompose into two, two and three constituent-feature
propositions, respectively. Twelve further propositions (X1--X12) cover features
that should not appear for the corresponding stage. Mention rate is defined in
Section~\ref{sec:metrics}.

Within-feature differences between correctly and incorrectly classified epochs
are assessed with one-sided subject-paired Wilcoxon signed-rank tests and
Holm--Bonferroni correction for multiple comparisons. All 11 required
propositions remain significant at the 0.05 level after correction, with
corrected $p < 1.2 \times 10^{-9}$ throughout and Cliff's $\delta \ge 0.88$.

By the audit design, an incorrectly classified rationale describes features of
the model's predicted (wrong) stage rather than features of the ground-truth
stage, against which it is audited. Part of each difference therefore reflects
this design choice in addition to perceptual hallucinations or contextual rule
misapplications by the model.

{\footnotesize
\setlength{\tabcolsep}{4pt}
\setlength{\LTleft}{0pt}\setlength{\LTright}{0pt}
\begin{longtable}{|L{0.048\textwidth}|L{0.058\textwidth}|L{0.386\textwidth}|L{0.180\textwidth}|L{0.238\textwidth}|}
\caption{Per-proposition results of the AASM-feature audit.}
\label{tab:props}\\
\hline
\textbf{ID} & \textbf{Stage} & \textbf{Proposition} & \textbf{Mention rate [95\% CI]} & \textbf{Correct $-$ incorrect $\Delta$ [pp, 95\% CI]} \\
\hline
\endfirsthead
\multicolumn{5}{l}{\footnotesize\itshape Table \thetable\ (continued)}\\
\hline
\textbf{ID} & \textbf{Stage} & \textbf{Proposition} & \textbf{Mention rate [95\% CI]} & \textbf{Correct $-$ incorrect $\Delta$ [pp, 95\% CI]} \\
\hline
\endhead
\hline
\endfoot
\hline
\endlastfoot
\multicolumn{5}{|l|}{\textit{Required propositions}}\\
\hline
F1 & W & Alpha rhythm (i.e., posterior dominant rhythm) over the occipital region for $>$50\% of the epoch & 81.5 [78.8, 83.8] & $+$90.8 [$+$89.7, $+$91.9] \\
F2 & W & 0.5--2 Hz eye blinks for $>$50\% of the epoch & 5.5 [4.8, 6.3] & $+$5.8 [$+$5.3, $+$6.4] \\
F3 & N1 & Attenuation or replacement of alpha rhythm by LAMF activity & 89.6 [88.0, 91.0] & $+$23.0 [$+$20.3, $+$25.8] \\
F4 & N1 & LAMF activity, or a 4--7 Hz background EEG & 89.4 [87.8, 90.8] & $+$23.2 [$+$20.3, $+$26.1] \\
F5 & N1 & Vertex sharp waves, or slow eye movements (SEMs) & 53.0 [50.6, 55.5] & $+$92.1 [$+$91.2, $+$93.1] \\
F6 & N2 & K complex unassociated with arousal & 60.0 [57.5, 62.4] & $+$72.8 [$+$70.3, $+$75.2] \\
F7 & N2 & Sleep spindle of 11--16 Hz with duration $\ge$0.5 s & 86.3 [84.8, 87.7] & $+$96.0 [$+$95.3, $+$96.6] \\
F8 & N3 & High-amplitude slow wave activity (peak-to-peak $>$75 \textmu{}V) for $\ge$20\% of the epoch & 81.8 [78.5, 84.5] & $+$99.0 [$+$98.2, $+$99.5] \\
F9 & R & LAMF EEG activity without K complexes or sleep spindles & 95.4 [94.7, 96.1] & $+$27.5 [$+$23.6, $+$31.3] \\
F10 & R & Low chin EMG tone & 86.7 [84.9, 88.4] & $+$65.7 [$+$62.0, $+$69.4] \\
F11 & R & Rapid eye movements (REMs) & 79.0 [76.1, 81.3] & $+$94.8 [$+$93.5, $+$95.9] \\
\hline
\multicolumn{5}{|l|}{\textit{Forbidden propositions}}\\
\hline
X1 & W & K complex unassociated with arousal & 1.5 [1.2, 2.1] & -- \\
X2 & W & Sleep spindle (11--16 Hz) & 1.7 [1.3, 2.2] & -- \\
X3 & W & High-amplitude slow wave activity for $\ge$20\% of the epoch & 1.5 [1.1, 2.0] & -- \\
X4 & N1 & K complex unassociated with arousal & 16.8 [15.0, 18.8] & -- \\
X5 & N1 & Sleep spindle & 23.6 [21.5, 25.7] & -- \\
X6 & N1 & Dominant alpha rhythm for $>$50\% of the epoch & 9.9 [8.5, 11.6] & -- \\
X7 & N2 & High-amplitude slow wave activity for $\ge$20\% of the epoch & 4.5 [3.6, 5.5] & -- \\
X8 & N2 & Both low chin EMG tone and REMs & 1.2 [1.0, 1.4] & -- \\
X9 & N3 & REMs & 0.1 [0.0, 0.2] & -- \\
X10 & N3 & Low chin EMG tone (e.g., low at REM level) & 32.7 [30.1, 35.1] & -- \\
X11 & R & K complex or sleep spindle & 4.2 [3.6, 4.9] & -- \\
X12 & R & High or elevated chin EMG tone & 1.3 [1.0, 1.7] & -- \\
\end{longtable}
\par}

\noindent
{\footnotesize
Values are averaged across the evaluation datasets. Per-dataset intervals are
obtained from 1,000 subject-level cluster bootstrap resamples and the averaged
interval is propagated from them. A correctness-stratified difference is not
defined for the forbidden propositions. LAMF, low-amplitude mixed-frequency;
SEMs, slow eye movements; REMs, rapid eye movements.
\par}

\paragraph{Judge prompts.}
Each rationale is audited independently by Qwen3.6-plus (accessed via Alibaba
Cloud DashScope) with temperature 0, max\_tokens 200, JSON-object response
format and thinking mode disabled. As in
Section~\ref{sec:prompts}, shaded boxes contain the verbatim prompt text
provided to the judge model and unshaded text provides editorial context.

\begin{promptbox}
You are an AASM v3 sleep-staging rule auditor. Given a piece of reasoning text
generated by a sleep-staging VLM (reasoning\_text, describing the model's
observations on the PSG waveform image) and the ground-truth sleep stage of the
sample, decide for each of the AASM feature propositions listed below whether
the reasoning text "affirmatively describes" that feature.

\textbf{\# Decision criterion (very important)}

For each proposition, output one of three values:
\begin{ptlist}
\item \textbf{"yes":} the reasoning text affirmatively asserts that the feature
      is present or that the condition is satisfied. Example: "alpha rhythm is
      present for 65\% of the epoch in O2-M1" $\rightarrow$ output yes for F1.
\item \textbf{"no":} the reasoning text explicitly asserts that the feature is
      absent / missing / replaced / excluded. Example: "no alpha rhythm
      visible", "alpha attenuated and replaced by LAMF", "absence of K complex"
      $\rightarrow$ output no for the corresponding presence-type proposition.
\item \textbf{"ambiguous":} the reasoning text does not mention the feature, or
      the wording is too vague to decide. Example: no mention of sleep spindle
      at all $\rightarrow$ output ambiguous for F7 / X2 / X5.
\end{ptlist}

\textbf{\# Recognition of negation}
\begin{ptlist}
\item "absence of X" / "no X observed" / "X is not present" $\rightarrow$ output
      no for an X-presence-type proposition.
\item "X is attenuated and replaced by Y" $\rightarrow$ output no for an
      X-presence-type proposition (X has been displaced).
\item "while X is present, it does not exceed 50\%" $\rightarrow$ output no for
      a "$>$50\% of the epoch" proposition (the quantitative threshold is not
      met).
\item \textbf{Quantitative thresholds:} F1 requires $>$50\% / F8 requires
      $\ge$20\% of the epoch with high amplitude / F2 requires $>$50\% / X3, X7
      require $\ge$20\%. When a proposition specifies a quantity, the
      quantitative condition must be met for yes; if the rationale gives only a
      qualitative description without a quantity, decide based on whether the
      rationale unambiguously affirms the condition.
\end{ptlist}

\textbf{\# Output format (mandatory JSON)}

Output a single JSON object whose keys are proposition IDs (e.g., "F3") and
whose values are "yes", "no", or "ambiguous". Do not output any additional text
and do not wrap the JSON in a Markdown code block.

Example output: \{"F3": "yes", "F4": "yes", "F5": "no", "X4": "ambiguous",
"X5": "ambiguous", "X6": "no"\}
\end{promptbox}

\noindent
In the user message, only the propositions triggered by the ground-truth stage
are listed. The placeholders \texttt{<STAGE>}, \texttt{<RATIONALE>} and
\texttt{<K>} are substituted at runtime, and each listed proposition is rendered
as a question using its verbatim wording from Table~\ref{tab:props}: F1, for
instance, becomes "Does the rationale affirmatively describe alpha rhythm
(i.e., posterior dominant rhythm) over the occipital region for $>$50\% of the
epoch?".

\begin{promptbox}
Ground-truth sleep stage: <STAGE>

Reasoning text (reasoning\_text):

"""

<RATIONALE>

"""

Please decide the following <K> propositions (output exactly one JSON object
whose keys are proposition IDs):
\begin{ptlist}
\item <ID> [REQ$|$FORBID]: <verbatim proposition wording>
\item ...
\end{ptlist}
\end{promptbox}

\section{Expert Ratings}
\label{sec:ratings}

\noindent
For each subject, the 10 rated epochs are drawn by stratified random sampling
across sleep stages using the largest-remainder method. Each rater works
independently.

{\footnotesize
\setlength{\LTleft}{0pt}\setlength{\LTright}{0pt}
\begin{longtable}{|L{0.155\textwidth}|L{0.791\textwidth}|}
\caption{Rating scale for expert evaluation of model reasoning.}
\label{tab:scale}\\
\hline
\textbf{Score} & \textbf{Description} \\
\hline
\endfirsthead
\multicolumn{2}{l}{\footnotesize\itshape Table \thetable\ (continued)}\\
\hline
\textbf{Score} & \textbf{Description} \\
\hline
\endhead
\hline
\endfoot
\hline
\endlastfoot
\multicolumn{2}{|p{0.957\textwidth}|}{\textbf{Factual Accuracy \& Perceptual Fidelity.} Whether the model's descriptions of waveform existence, frequency, amplitude, and temporal proportion faithfully reflect the rendered PSG image.}\\
\hline
5 (Excellent) & All waveform descriptions match the image with zero factual errors in existence, frequency, amplitude, or epoch proportion. \\
4 (Good) & Core stage-defining features accurately described; only minor deviations in non-critical background signals. \\
3 (Acceptable) & Waveform existence correctly identified; localized attribute or temporal estimation errors that do not alter the overall staging determination. \\
2 (Marginal) & Significant attribute errors, or key waveform amplitude/frequency estimates severely violate the rendering scales. \\
1 (Poor) & Fabricated waveforms described (absent in image), or the majority of feature descriptions contradict image evidence. \\
0 (Fail) & Descriptions entirely unrelated to the PSG image. \\
\hline
\multicolumn{2}{|p{0.957\textwidth}|}{\textbf{Diagnostic Evidence Comprehensiveness.} Whether the model systematically presents multi-channel evidence (EEG, EOG, EMG), stage-defining features, exclusionary features, and adjacent-epoch context required by AASM rules.}\\
\hline
5 (Excellent) & All three signal systems covered; stage-defining features exhaustive; exclusionary evidence depth proportionate to staging ambiguity; adjacent-epoch context utilized when applicable. \\
4 (Good) & Core channels and key stage-defining features complete; minor brevity in adjacent-epoch utilization or secondary exclusionary features, not compromising the evidence chain. \\
3 (Acceptable) & Core stage-defining features provided but with one of: a non-key channel omitted, insufficient exclusionary evidence for an ambiguous epoch, or adjacent-epoch context underutilized. \\
2 (Marginal) & Key channel descriptions omitted per AASM requirements, or necessary exclusionary evidence absent in a highly confusable epoch. \\
1 (Poor) & Only single-channel description or conclusory judgment; no substantive multi-channel feature analysis. \\
0 (Fail) & No channel-feature evidence usable for clinical interpretation. \\
\hline
\multicolumn{2}{|p{0.957\textwidth}|}{\textbf{Logical Coherence \& Guideline Concordance.} Whether the reasoning chain from observed features to the staging conclusion is rigorous, self-consistent, and adherent to the AASM scoring manual.}\\
\hline
5 (Excellent) & Rigorous causal logic from evidence to conclusion; cited AASM rules precisely applicable; reasoning forms a complete logical closed-loop at expert level. \\
4 (Good) & Core reasoning correct and evidence sufficiently supports the conclusion; AASM rules applied accurately; minor inefficiencies in secondary steps only. \\
3 (Acceptable) & Correct conclusion with accurate core reasoning; slight logical leaps or loose integration between cited rules and described evidence. \\
2 (Marginal) & Internal contradictions present; described evidence mismatches the cited scoring rules or reasoning direction, failing to support the conclusion. \\
1 (Poor) & Fundamental opposition between the factual evidence presented in the reasoning and the final staging conclusion. \\
0 (Fail) & No coherent reasoning present; no valid logical chain formed. \\
\end{longtable}
\par}

\noindent
{\footnotesize
Each dimension is scored independently on a 0--5 integer scale. AASM, American
Academy of Sleep Medicine.
\par}

\paragraph{Per-rater score distributions.}
Table~\ref{tab:raterdist} reports the two raters separately over all 5,400 rated
samples; the counts averaged over them are the ones plotted in Figure~3 of the
main paper.

{\footnotesize
\setlength{\tabcolsep}{5pt}
\setlength{\LTleft}{0pt}\setlength{\LTright}{0pt}
\begin{longtable}{@{\extracolsep{\fill}}l c c rrrrrr@{}}
\caption{Per-rater score distributions over the 5,400 rated samples.}
\label{tab:raterdist}\\
\toprule
\textbf{Dimension} & \textbf{Rater} & \textbf{Mean} & \textbf{0} & \textbf{1} & \textbf{2} & \textbf{3} & \textbf{4} & \textbf{5} \\
\midrule
\endfirsthead
\toprule
\textbf{Dimension} & \textbf{Rater} & \textbf{Mean} & \textbf{0} & \textbf{1} & \textbf{2} & \textbf{3} & \textbf{4} & \textbf{5} \\
\midrule
\endhead
\midrule
\endfoot
\bottomrule
\endlastfoot
\multirow{2}{*}{Factual accuracy} & 1 & 4.14 & 3 & 434 & 152 & 240 & 1,938 & 2,633 \\
 & 2 & 3.47 & 25 & 395 & 561 & 1,168 & 2,552 & 699 \\
\midrule
\multirow{2}{*}{Evidence comprehensiveness} & 1 & 4.06 & 4 & 434 & 169 & 159 & 2,516 & 2,118 \\
 & 2 & 3.65 & 25 & 360 & 547 & 918 & 2,262 & 1,288 \\
\midrule
\multirow{2}{*}{Logical coherence} & 1 & 4.08 & 4 & 568 & 159 & 120 & 1,976 & 2,573 \\
 & 2 & 3.70 & 25 & 360 & 565 & 918 & 1,916 & 1,616 \\
\end{longtable}
\par}

\paragraph{Inter-rater reliability.}
Pooled over all paired ratings, 78.0\% of the paired scores differ by at most
one point, the quadratic-weighted Cohen's $\kappa$ is 0.496, and the ICC(2,k) is
0.663 (95\% CI 0.306--0.806).

\section{Additional Qualitative Examples}
\label{sec:qual}

\noindent
Each figure below shows one scored epoch: three consecutive 30-second PSG epoch
images (preceding, current, subsequent) with six channels (F4-M1, C4-M1, O2-M1,
LOC, ROC, Chin EMG), followed by the stage predicted by SleepVLM, the AASM rule
identifiers it cites and its rationale. The stage printed above the middle image
is the ground-truth label. A check mark denotes a correct classification and a
cross a misclassification; each misclassified example carries an expert note
identifying the applicable rule. The cited rules are defined in
Table~\ref{tab:rules}. Rationales are condensed for brevity without altering
their substantive meaning.

\subsection{Correctly Classified Epochs}

\begin{figure}[H]
  \centering
  \includegraphics[width=\linewidth]{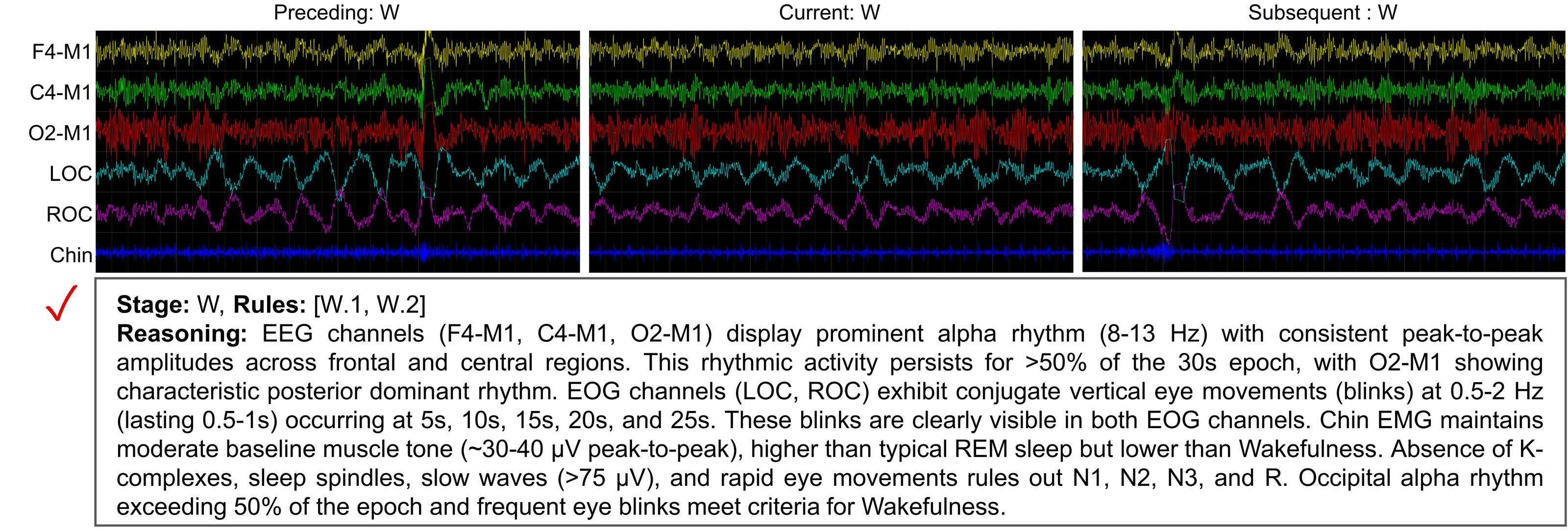}
  \caption{Stage W (Rules W.1, W.2).}
  \label{fig:se1}
\end{figure}

\begin{figure}[H]
  \centering
  \includegraphics[width=\linewidth]{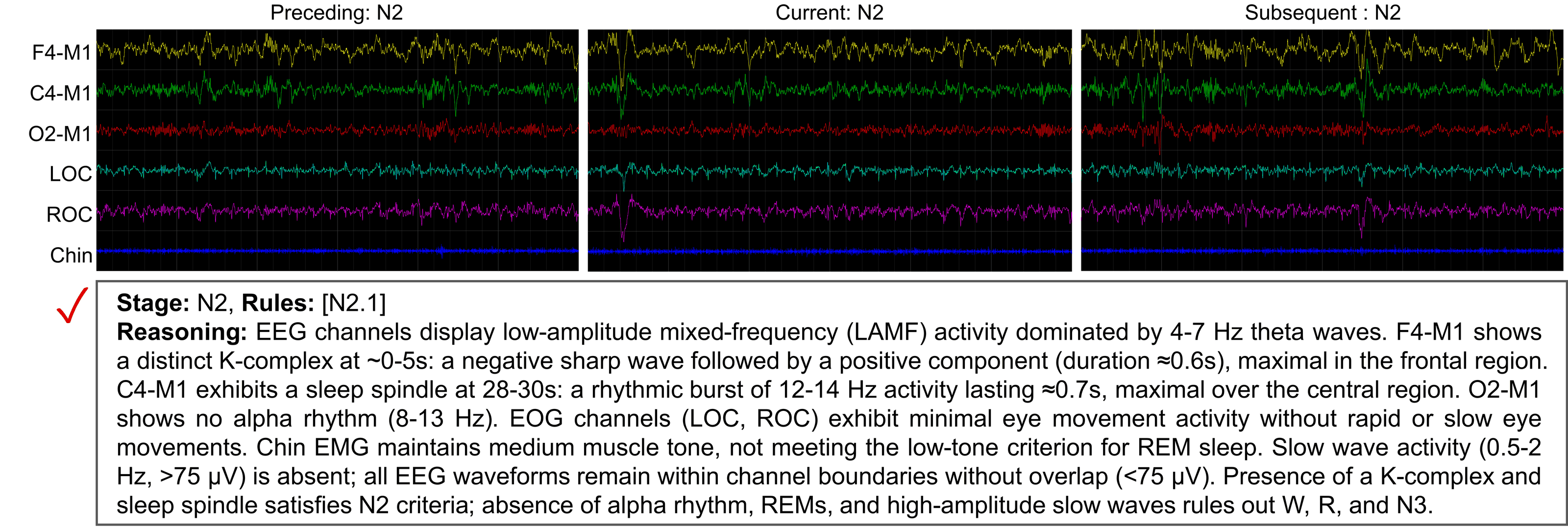}
  \caption{Stage N2 (Rule N2.1).}
  \label{fig:se2}
\end{figure}

\begin{figure}[H]
  \centering
  \includegraphics[width=\linewidth]{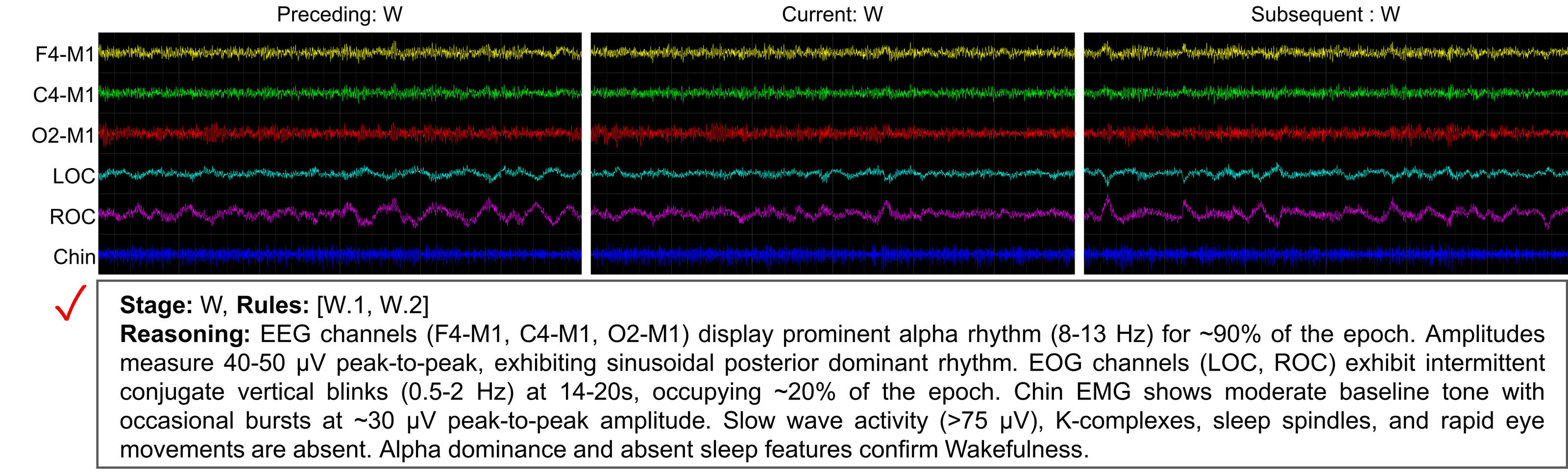}
  \caption{Stage W (Rules W.1, W.2).}
  \label{fig:se3}
\end{figure}

\begin{figure}[H]
  \centering
  \includegraphics[width=\linewidth]{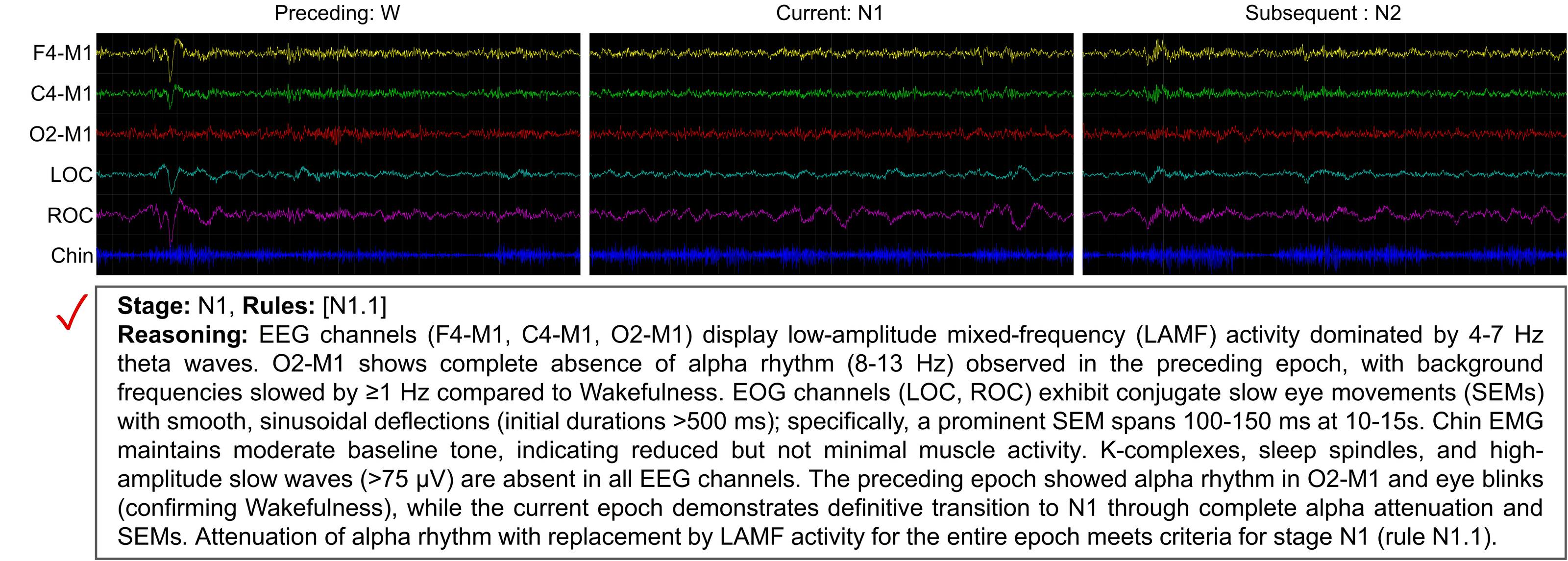}
  \caption{Stage N1 (Rule N1.1).}
  \label{fig:se4}
\end{figure}

\begin{figure}[H]
  \centering
  \includegraphics[width=\linewidth]{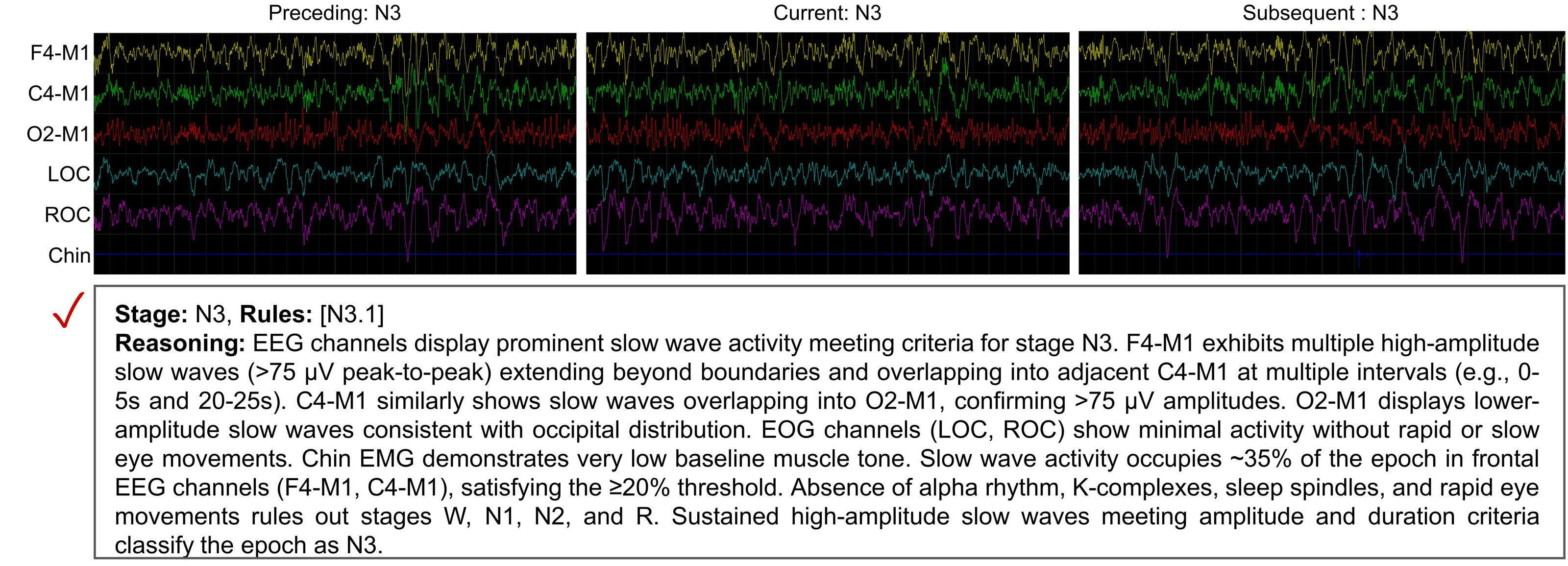}
  \caption{Stage N3 (Rule N3.1).}
  \label{fig:se6}
\end{figure}

\begin{figure}[H]
  \centering
  \includegraphics[width=\linewidth]{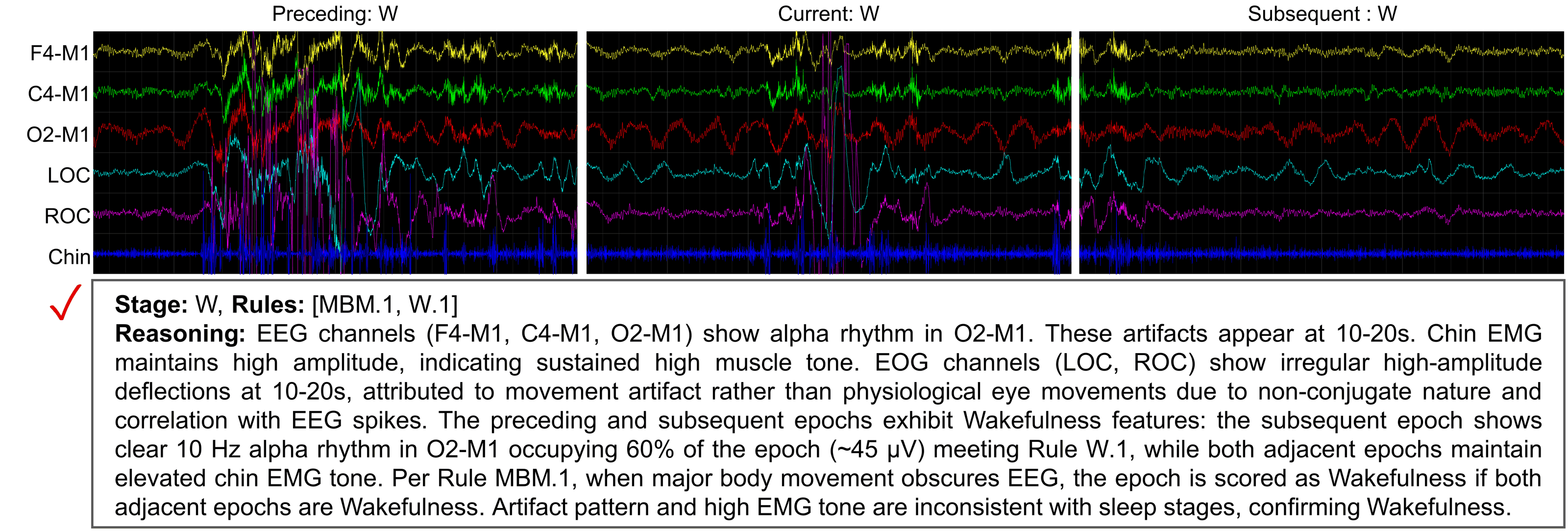}
  \caption{Stage W (Rules MBM.1, W.1).}
  \label{fig:se7}
\end{figure}

\begin{figure}[H]
  \centering
  \includegraphics[width=\linewidth]{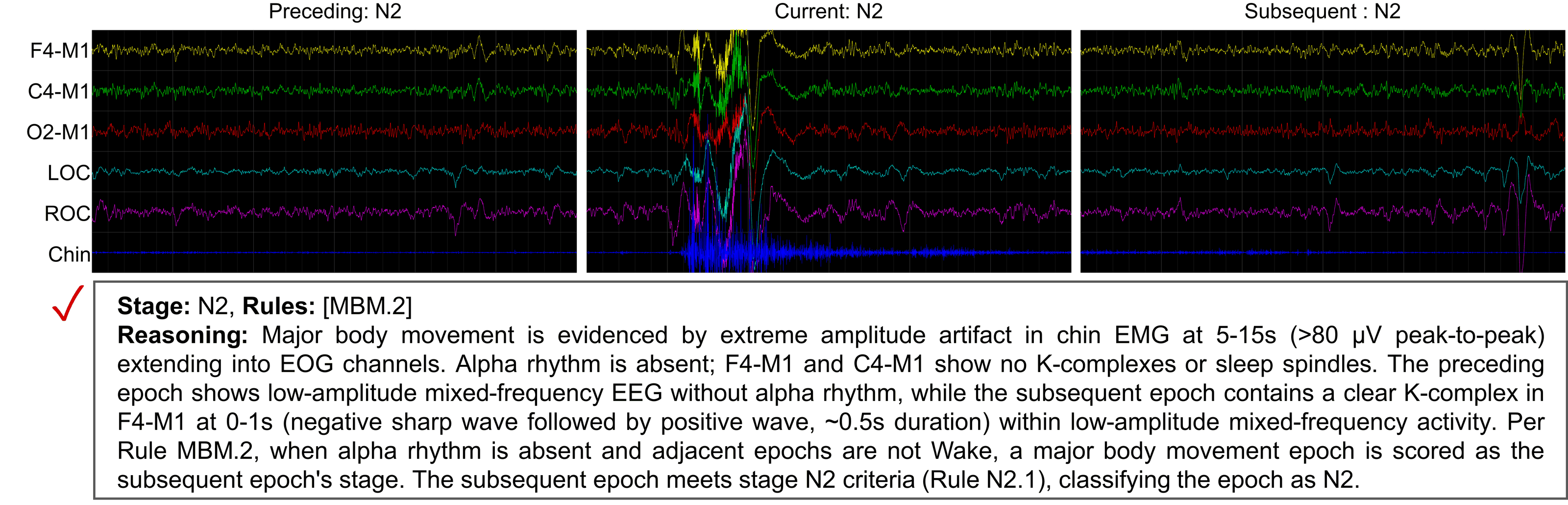}
  \caption{Stage N2 (Rule MBM.2).}
  \label{fig:se9}
\end{figure}

\begin{figure}[H]
  \centering
  \includegraphics[width=\linewidth]{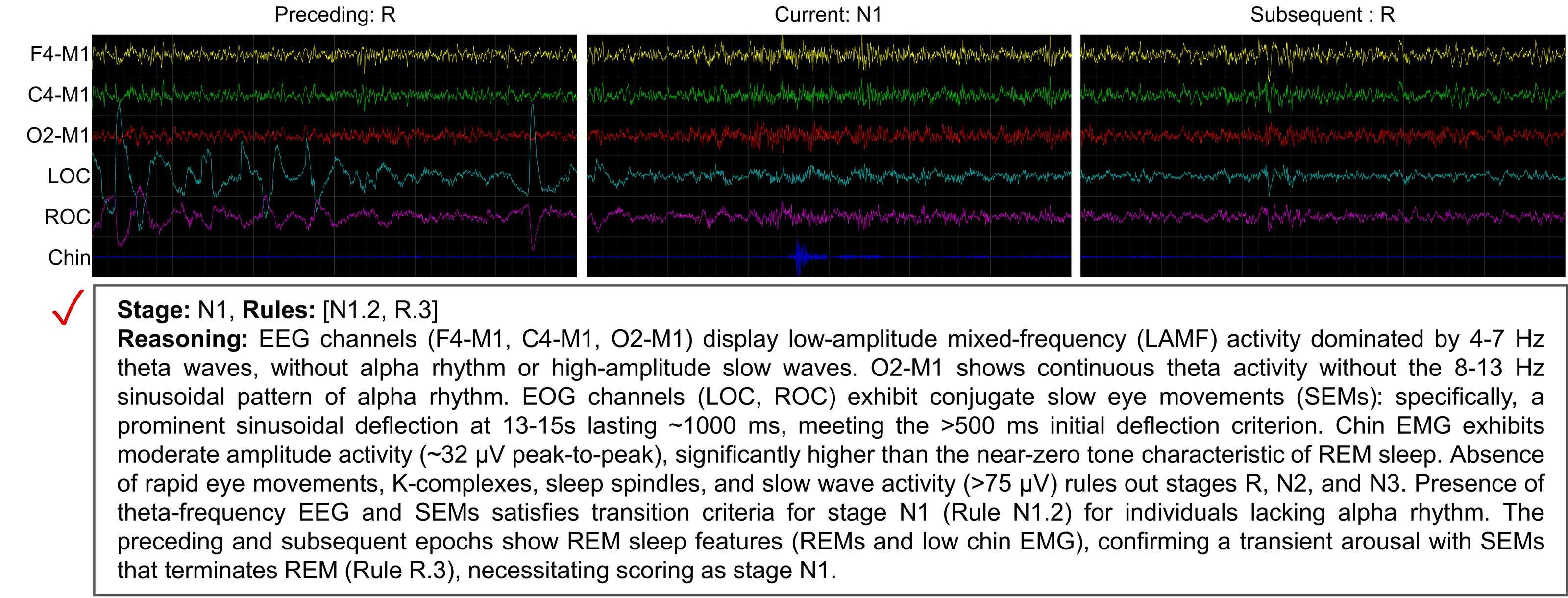}
  \caption{Stage N1 (Rules N1.2, R.3).}
  \label{fig:se10}
\end{figure}

\clearpage
\subsection{Misclassified Epochs}

\begin{figure}[H]
  \centering
  \includegraphics[width=\linewidth]{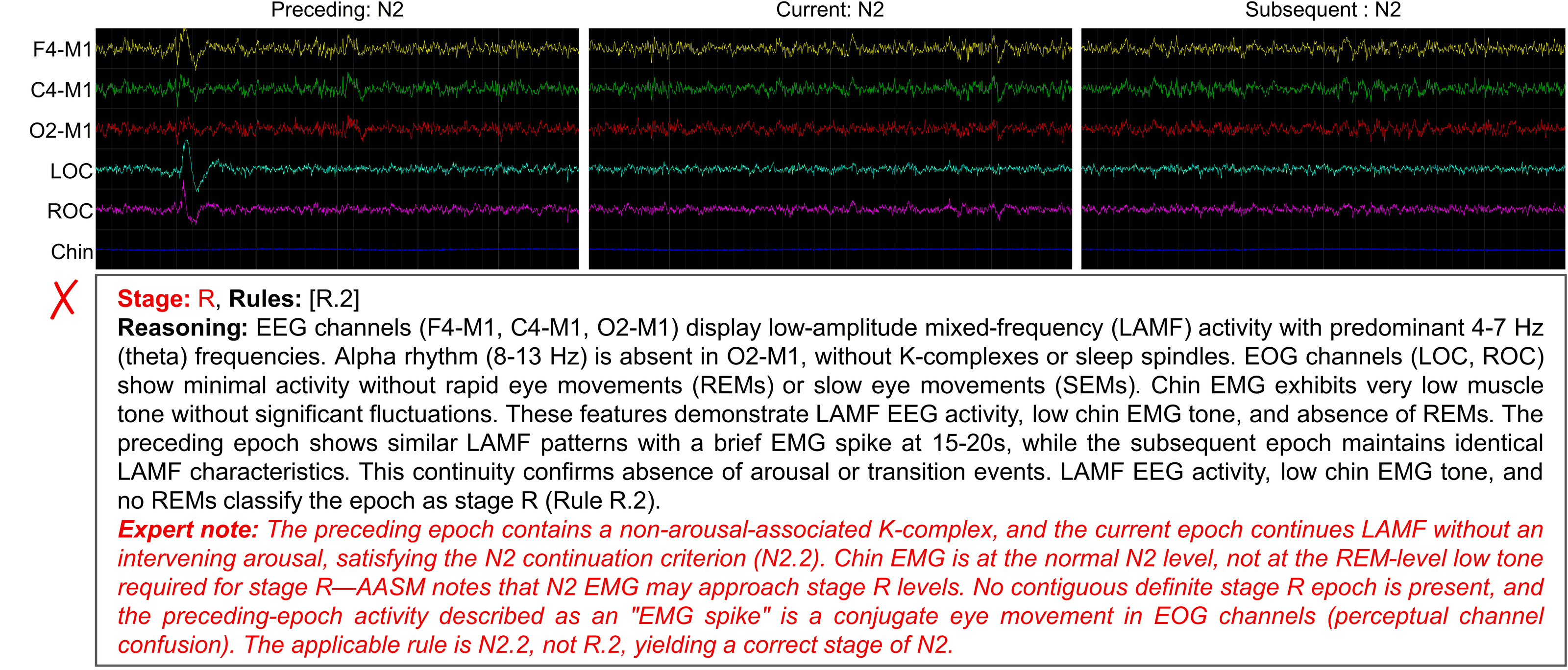}
  \caption{Ground-truth stage N2 (Rule N2.2); predicted stage R (Rule R.2).}
  \label{fig:se5}
\end{figure}

\begin{figure}[H]
  \centering
  \includegraphics[width=\linewidth]{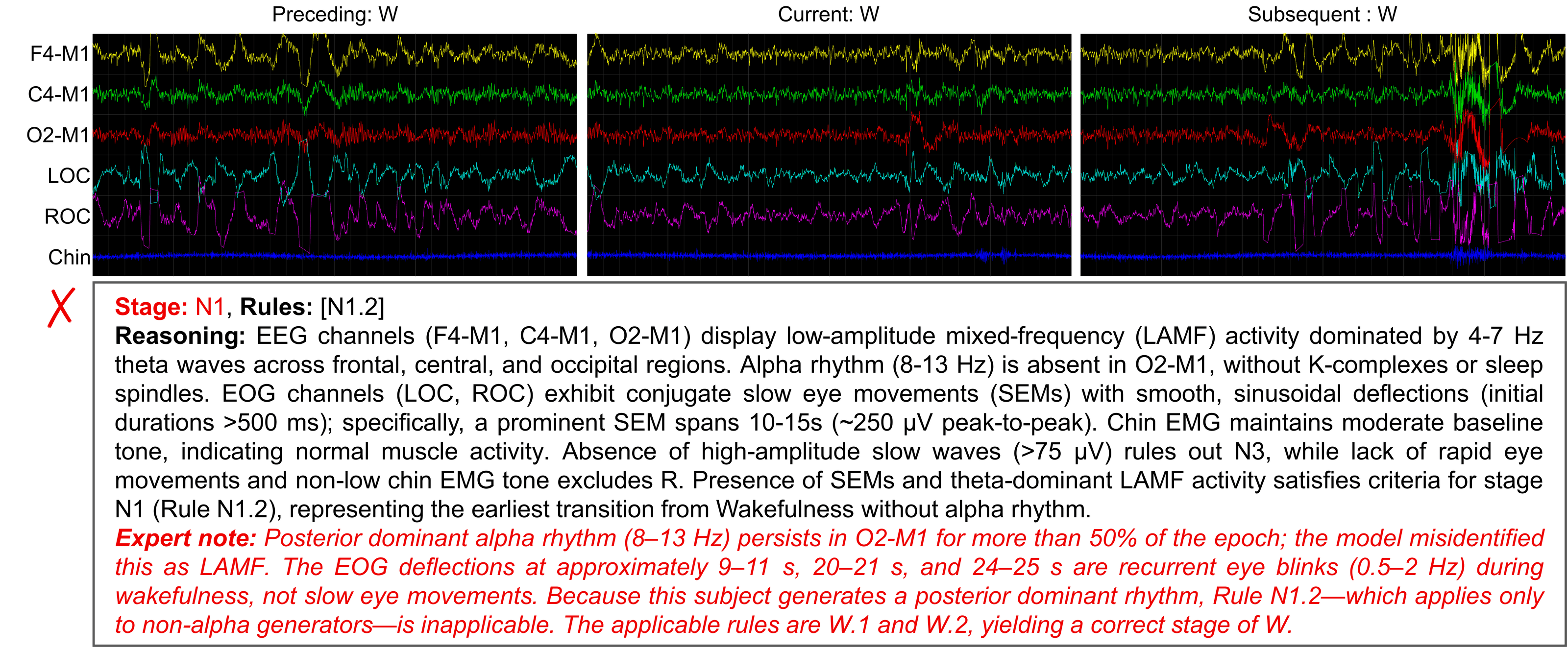}
  \caption{Ground-truth stage W (Rules W.1, W.2); predicted stage N1
    (Rule N1.2).}
  \label{fig:se8}
\end{figure}

\bibliographystyle{aaai2027}
\bibliography{references}